\RequirePackage{silence}
\documentclass[runningheads]{llncs}

\usepackage[mobile]{eccv}

\usepackage{eccvabbrv}

\usepackage{graphicx}
\usepackage{booktabs}

\usepackage[accsupp]{axessibility}  %

\usepackage{algorithm}
\usepackage{algpseudocodex}
\usepackage{multirow}
\usepackage{array}
\usepackage{amsmath}
\usepackage{wrapfig}
\usepackage{longfigure}
\usepackage{placeins}
\usepackage{needspace}
\usepackage{nicefrac}
\usepackage[dvipsnames]{xcolor}
\usepackage{colortbl}
\usepackage{rotating}
\usepackage{pgf}
\usepackage{tikz}
\usetikzlibrary{arrows.meta,backgrounds,calc,fit,positioning}
\usepackage{pifont}

\definecolor{joltLatent}{RGB}{63,63,193}
\definecolor{joltImage}{RGB}{255,192,0}
\definecolor{joltPrompt}{RGB}{0,140,60}
\definecolor{joltLRtoHR}{RGB}{0,117,120}
\definecolor{joltHRtoLR}{RGB}{176,82,0}
\colorlet{joltFeedback}{joltLRtoHR}

\DeclareMathSizes{5.6}{5.6}{5}{5}

\newcommand{\METHODNAME}{JoLT}

\newcommand{\PreserveBackslash}[1]{\let\temp=\\#1\let\\=\temp}
\newcolumntype{C}[1]{>{\PreserveBackslash\centering}p{#1}}
\newcolumntype{R}[1]{>{\PreserveBackslash\raggedleft}p{#1}}
\newcolumntype{L}[1]{>{\PreserveBackslash\raggedright}p{#1}}
\allowdisplaybreaks

\usepackage[breaklinks,colorlinks,citecolor=eccvblue,linkcolor=eccvblue,urlcolor=eccvblue]{hyperref}

\usepackage{orcidlink}

\begin{document}

\title{JoLT: Joint Latent Trajectories for Context-Guided High-Resolution Tiled Generation}

\titlerunning{JoLT: Joint Latent Trajectories}

\author{
  Mathis Koroglu\inst{1,2}\orcidlink{0009-0000-6589-3151} \and
  Guillaume Jeanneret\inst{2}\orcidlink{0000-0002-0055-7816} \and
  Hugo Caselles-Dupré\inst{1}\orcidlink{0000-0003-2711-5732} \and
  Matthieu Cord\inst{2,3}\orcidlink{0000-0002-0627-5844} \and
  Arnaud Dapogny\inst{2}
}

\authorrunning{M.~Koroglu \etal}

\institute{
  Obvious Research, Paris, France\\
  \and
  Sorbonne Université, CNRS, ISIR, F-75005 Paris, France\\
  \and
  Valeo.ai, Paris, France
}

\maketitle

\begin{abstract}
  Although text-to-image generative models produce impressive results, they struggle to generate densely detailed, high-resolution (HR) images. Current literature addresses this issue with a low-to-high-resolution approach. First, a low-resolution (LR) image is generated. Then, an upsampled version is generated using the LR image as an additional cue. In this paper, we present Joint Latent Trajectories (\METHODNAME). To generate an image, \METHODNAME{} uses two streams that jointly denoise LR and HR latent images at each sampling step. The LR latent controls the overall layout, while the HR latent controls the details. We interconnect both branches to jointly integrate their information. We extensively validate our method, demonstrating its advantages over competing baselines. The resulting images are not only richly detailed but also visually pleasing, opening new avenues for artistic creation.

\end{abstract}

\begin{figure*}[t]
  \centering
  \resizebox{\textwidth}{!}{%
    \begin{tikzpicture}[x=1cm,y=1cm]
      \tikzset{
        main panel/.style={draw=black!24,line width=.55pt},
        input panel/.style={rounded corners=1.2mm,draw=black!22,
          fill=black!2,line width=.55pt},
        input card/.style={rounded corners=1mm,draw=#1!70!black,
          fill=#1!7,line width=.65pt,minimum width=4.18cm,
          minimum height=.66cm,align=center,inner xsep=4pt,inner ysep=2pt,
          font=\sffamily\small},
        input title/.style={font=\sffamily\small\bfseries},
        master flow/.style={-{Stealth[length=2.6mm,width=1.8mm]},
          draw=joltPrompt!72!black,line width=.95pt},
        refinement flow/.style={-{Stealth[length=2.6mm,width=1.8mm]},
          draw=joltImage!76!black,line width=.95pt},
        complexity flow/.style={-{Stealth[length=2.6mm,width=1.8mm]},
          draw=joltFeedback!76!black,line width=.95pt},
        crop guide/.style={draw=white,line width=.8pt,
          dash pattern=on 3pt off 2pt,
          preaction={draw=black!62,line width=1.65pt}},
        method label/.style={anchor=south west,rounded corners=.7mm,
          fill=black,fill opacity=.52,text opacity=1,text=white,
          inner xsep=5pt,inner ysep=3pt},
        resolution label/.style={method label,rounded corners=.5mm,
          inner xsep=4pt,inner ysep=1pt},
        capability band/.style={rounded corners=.8mm,draw=black!23,
          fill=black!3,line width=.55pt},
      }

      \path[use as bounding box] (-.08,-.08) rectangle (13.75,9.90);

      \def\fluxCropCenterX{0.285}
      \def\fluxCropCenterY{0.52}
      \def\joltCropCenterX{0.155}
      \def\joltCropCenterY{0.50}

      \node[anchor=south west,inner sep=0] (fluximage) at (0,1.00)
        {\includegraphics[width=4.41cm,height=4.41cm]{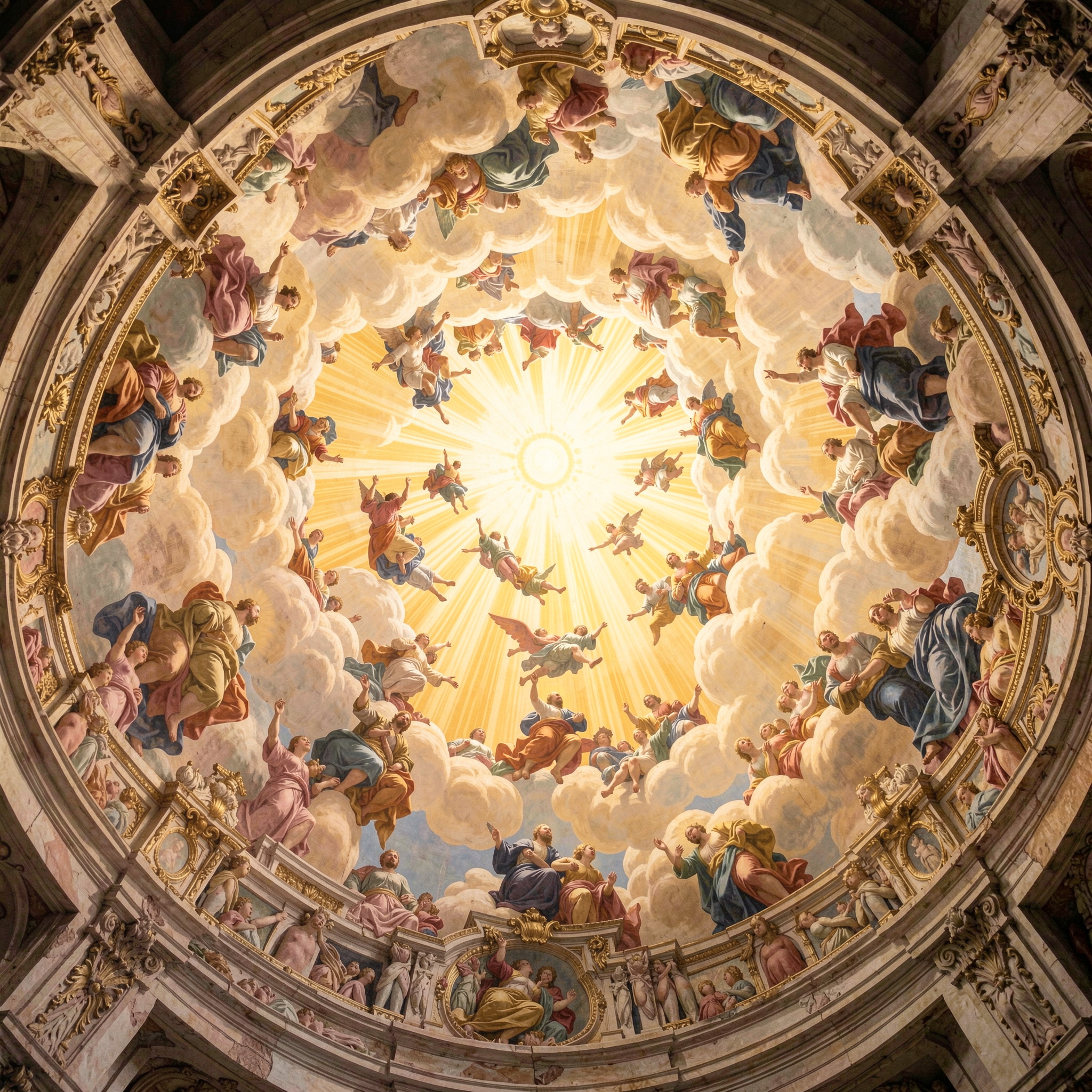}};
      \draw[main panel] (0,1.00) rectangle (4.41,5.41);

      \pgfmathsetmacro{\fluxCropLeftTrim}{max(0,min(1824,
        2048*\fluxCropCenterX-112))}
      \pgfmathsetmacro{\fluxCropTopTrim}{max(0,min(1824,
        2048*\fluxCropCenterY-112))}
      \pgfmathsetmacro{\fluxCropRightTrim}{1824-\fluxCropLeftTrim}
      \pgfmathsetmacro{\fluxCropBottomTrim}{1824-\fluxCropTopTrim}
      \pgfmathsetmacro{\fluxCropXLeft}{(4.41/2048)*\fluxCropLeftTrim}
      \pgfmathsetmacro{\fluxCropXRight}{(4.41/2048)*(\fluxCropLeftTrim+224)}
      \pgfmathsetmacro{\fluxCropYBottom}{1+(4.41/2048)*\fluxCropBottomTrim}
      \pgfmathsetmacro{\fluxCropYTop}{1+(4.41/2048)*(\fluxCropBottomTrim+224)}
      \coordinate (fluxCropSW) at (\fluxCropXLeft,\fluxCropYBottom);
      \coordinate (fluxCropNW) at (\fluxCropXLeft,\fluxCropYTop);
      \coordinate (fluxCropNE) at (\fluxCropXRight,\fluxCropYTop);
      \coordinate (fluxCropSE) at (\fluxCropXRight,\fluxCropYBottom);
      \draw[crop guide] (fluxCropSW) rectangle (fluxCropNE);
      \node[anchor=south west,inner sep=0,draw=white,line width=1.35pt]
        (fluxzoom) at (2.24,3.22)
        {\includegraphics[width=1.95cm,
          trim=\fluxCropLeftTrim bp \fluxCropBottomTrim bp
            \fluxCropRightTrim bp \fluxCropTopTrim bp,clip]
          {figures/main_teaser/base_generation_flux2_klein_9b_2048.jpg}};
      \draw[crop guide] (fluxCropNW) -- (fluxzoom.north west);
      \draw[crop guide] (fluxCropSE) -- (fluxzoom.south east);
      \node[resolution label,font=\sffamily\scriptsize\mdseries]
        (fluxresolution) at (.14,1.08) {$2048^2$ px};
      \node[method label,font=\sffamily\large\bfseries]
        at ([yshift=.4mm]fluxresolution.north west)
        {FLUX.2 Klein-9B};

      \draw[input panel] (0,5.61) rectangle (4.41,9.82);
      \node[anchor=west,font=\sffamily\small\bfseries,text=black!68]
        at (.28,9.48) {INPUTS};
      \node[input card=joltFeedback,minimum height=.76cm]
        (alpha) at (2.205,8.73) {};
      \node[input title] at
        ([yshift=1.6mm]alpha.center) {Complexity strength};
      \node[font=\sffamily\small] at
        ([yshift=-1.6mm]alpha.center) {$\alpha\in[0,1]$};
      \node[input card=joltImage,minimum height=.94cm,
        below=1.6mm of alpha] (refinement) {};
      \node[input title] at
        ([yshift=1.8mm]refinement.center) {Detail prompt $\bar{c}$};
      \node[font=\sffamily\scriptsize\itshape] at
        ([yshift=-1.8mm]refinement.center) {Enhance image, add details};
      \node[input card=joltPrompt,minimum height=1.20cm,
        below=1.6mm of refinement] (master) {};
      \node[input title] at
        ([yshift=3.5mm]master.center) {Master prompt $c$};
      \node[font=\sffamily\tiny\itshape,align=center] at
        ([yshift=-1.5mm]master.center)
        {A soaring painted dome \ldots\\[-.1ex]
         angels ascending into golden light,\\[-.1ex]
         a luminous Baroque ceiling fresco.};

      \draw[master flow] (master.south) -- (2.205,5.41);

      \node[anchor=south west,inner sep=0] (joltimage) at (4.85,1.00)
        {\includegraphics[width=8.82cm,height=8.82cm]
          {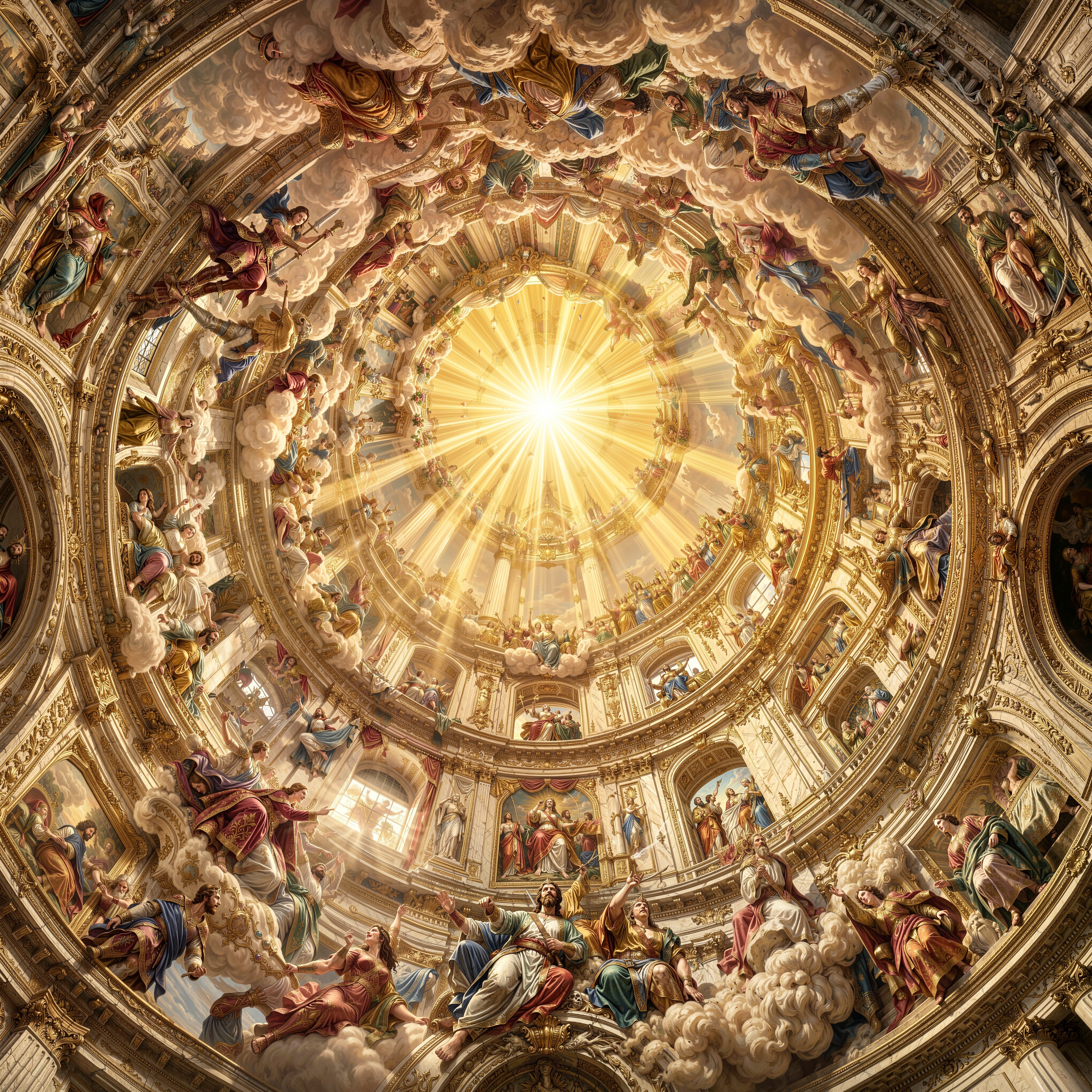}};
      \draw[main panel] (4.85,1.00) rectangle (13.67,9.82);

      \draw[complexity flow] (alpha.east) -- (joltimage.west |- alpha.east);
      \draw[refinement flow] (refinement.east) --
        (joltimage.west |- refinement.east);
      \draw[master flow] (master.east) -- (joltimage.west |- master.east);

      \pgfmathsetmacro{\joltCropLeftTrim}{max(0,min(3646,
        4096*\joltCropCenterX-225))}
      \pgfmathsetmacro{\joltCropTopTrim}{max(0,min(3646,
        4096*\joltCropCenterY-225))}
      \pgfmathsetmacro{\joltCropRightTrim}{3646-\joltCropLeftTrim}
      \pgfmathsetmacro{\joltCropBottomTrim}{3646-\joltCropTopTrim}
      \pgfmathsetmacro{\joltCropXLeft}{4.85+(8.82/4096)*\joltCropLeftTrim}
      \pgfmathsetmacro{\joltCropXRight}{4.85+(8.82/4096)*(\joltCropLeftTrim+450)}
      \pgfmathsetmacro{\joltCropYBottom}{1+(8.82/4096)*\joltCropBottomTrim}
      \pgfmathsetmacro{\joltCropYTop}{1+(8.82/4096)*(\joltCropBottomTrim+450)}
      \coordinate (joltCropSW) at (\joltCropXLeft,\joltCropYBottom);
      \coordinate (joltCropNW) at (\joltCropXLeft,\joltCropYTop);
      \coordinate (joltCropNE) at (\joltCropXRight,\joltCropYTop);
      \coordinate (joltCropSE) at (\joltCropXRight,\joltCropYBottom);
      \draw[crop guide] (joltCropSW) rectangle (joltCropNE);
      \node[anchor=south west,inner sep=0,draw=white,line width=1.55pt]
        (joltzoom) at (9.77,5.94)
        {\includegraphics[width=3.65cm,
          trim=\joltCropLeftTrim bp \joltCropBottomTrim bp
            \joltCropRightTrim bp \joltCropTopTrim bp,clip]
          {figures/main_teaser/jolt_refined_output_4096.jpg}};
      \draw[crop guide] (joltCropNW) -- (joltzoom.north west);
      \draw[crop guide] (joltCropSE) -- (joltzoom.south east);
      \node[resolution label,font=\fontfamily{phv}\selectfont\scriptsize\mdseries]
        (joltresolution) at (5.10,1.08) {$4096^2$ px};
      \node[method label,font=\fontfamily{phv}\selectfont\LARGE\mdseries]
        at ([yshift=.4mm]joltresolution.north west)
        {\METHODNAME{}};

      \draw[capability band,fill=red!3] (0,0) rectangle (4.41,.82);
      \node[font=\sffamily\scriptsize\bfseries,text=black!78] at (2.205,.41)
        {\textcolor{red!78!black}{\ding{55}}\; Simpler visuals up to 2k};

      \draw[capability band,fill=joltPrompt!4] (4.85,0) rectangle (13.67,.82);
      \node[font=\sffamily\scriptsize\bfseries,text=black!78] at (6.32,.41)
        {\textcolor{joltPrompt!78!black}{\ding{51}}\; Detail control};
      \node[font=\sffamily\scriptsize\bfseries,text=black!78] at (9.26,.41)
        {\textcolor{joltPrompt!78!black}{\ding{51}}\; Unbounded Resolution};
      \node[font=\sffamily\scriptsize\bfseries,text=black!78] at (12.20,.41)
        {\textcolor{joltPrompt!78!black}{\ding{51}}\; High Quality};
    \end{tikzpicture}%
  }
  \caption{\textbf{Controllable high-resolution generation with \METHODNAME.}
  The master prompt $c$ conditions both the FLUX.2 Klein-9B base generation (left) and
  our result (right), while the detail prompt $\bar{c}$ and complexity
  strength $\alpha$ control the added local detail. Dashed boxes identify matched
  groups of figures near the image center, enlarged in the overlaid crops. The base image is displayed
  at half the width and height of our result; the bands summarize their qualitative
  capabilities.}
  \label{fig:main}
\end{figure*}

\section{Introduction}
\label{sec:intro}

The introduction of novel generative technologies, such as text-to-image (T2I) models, has sparked interest in adopting these tools for artistic creation. However, using vanilla T2I models can still feel limited, as even the best-performing models, such as FLUX.2, produce relatively little detail. This shortfall is most consequential for large-format works meant for physical display, such as prints or murals, which are read at two distances: the composition from afar, the fine detail up close. Artists would like to freely control intricate local details, including through natural language, without sacrificing the coherence and aesthetics of the global composition.

To set a frame of reference, let us clearly state what we refer to as a ``complex'' or ``detailed'' image. We consider an image to be complex or detailed if the density of semantic information is high locally, yet the scene can be understood globally. However, it is not complex if it only contains texturized or recurring patterns. For example, increasing the sharpness of an image is perceived by the human eye as more detailed, yet it does not fit into our definition. \cref{fig:main} shows an example of a generation made with FLUX.2 Klein-9B and another made with our proposed approach, both using the same input prompt. As can be seen here, FLUX.2 Klein-9B's generation is simple, even with a detailed prompt. In contrast, our image is locally complex yet globally coherent.

Current training-free methods typically generate large, aesthetically pleasing images in two steps. First, they create an LR image and, second, they upsample it to generate details or refine textures.
This strategy has two drawbacks. First, since the LR image is bound by the inherent complexity limitations of the underlying model, the refining step barely enhances the richness of the textures. Second, since the LR image has already been generated, it cannot be influenced by HR information, making it difficult to integrate fine-grained elements seamlessly into the global layout.
Existing approaches therefore face an inherent trade-off between detail and global coherence. This motivates a different strategy for generating these images.

\begingroup\hfuzz=1.5pt
In this paper, we focus on generating densely detailed HR images in a training-free manner. Instead of using the conventional low-to-high-resolution generation paradigm, we propose a novel two-stream approach that jointly generates LR and HR latent features. Our method, called Joint Latent Trajectories (\METHODNAME), transfers information from the LR stream to the HR stream via the context input of FLUX.2, then feeds the evolving HR representation back into the LR stream. Consequently, the detail prompt does more than populate local regions: together with the master prompt, it helps shape the style and rendering of the final result. This bidirectional exchange integrates both resolutions into a coherent scene while giving users control over image complexity and stylization.
\par\endgroup

To summarize, our contributions are as follows:

\begin{itemize}
    \item A Novel Joint-Latent Paradigm: We propose \METHODNAME{}, a training-free pipeline for generating high-resolution, complex images. \METHODNAME{} transforms the standard low-to-high generation paradigm into a simultaneous low-and-high generation paradigm.
    \item Bidirectional Latent Information Exchange: To transfer information between the HR and LR branches, we propose a cross-resolution feedback mechanism that leverages the in-context capabilities of FLUX.2 and variance-matched latent subsampling.
    \item Controllable High-Density Synthesis: We enable users to explicitly control visual complexity ($\alpha$), as well as the style and local content of the final result through the detail prompt ($\bar{c}$), without requiring fine-tuning.
\end{itemize}

With a series of strong empirical studies, we show the advantages of \METHODNAME{} over recent methods and baselines.

\paragraph{Project resources.}
The project page, source code, and evaluation prompts are available at \url{https://obvious-research.github.io/jolt/}.

\section{Related Work}
\label{sec:related}

\paragraph{Diffusion and rectified-flow generative backbones.}
Modern image generation is largely built on diffusion and score-based models that reverse a gradual noising process~\cite{ho2020ddpm,song2021scoresde}, with samplers such as DDIM~\cite{song2021ddim} and classifier-free guidance~\cite{ho2021classifierfree}. Since pixel-space denoising scales poorly with resolution, latent diffusion moves the process into a compressed autoencoder space and makes high-resolution text-to-image synthesis practical~\cite{rombach2022high}. Newer systems shift toward transformer backbones and rectified-flow objectives that reshape the generative trajectory: Diffusion Transformers~\cite{peebles2023dit} denoise over latent patches, making capacity and token resolution central scaling variables, Stable Diffusion~3~\cite{esser2024scaling} couples a multimodal diffusion transformer with a rectified-flow formulation, and families such as FLUX~\cite{flux2024} and HunyuanImage~\cite{cao2025hunyuanimage} carry the trend into large-scale democratization in the creative community. %
A distinct paradigm is image-conditioned, in-context generation, where a reference image is optionally fed as additional conditioning. In-Context LoRA~\cite{huang2024incontextlora} adapts a diffusion transformer to jointly generate related images from concatenated in-context examples, and OminiControl~\cite{tan2024ominicontrol} injects image conditions by appending their tokens to the generation sequence with minimal added parameters. FLUX.1 Kontext~\cite{labs2025flux1kontextflowmatching} unifies generation and editing in one latent flow-matching model. Our method builds on this in-context view, conditioning on a lower-resolution image as it is being denoised.

\paragraph{High-detail and high-resolution generative synthesis.}
High-resolution generative synthesis attempts to create large, coherent, semantically rich images from text and/or reference context, outside the generator's native training resolution.
MultiDiffusion~\cite{bar2023multidiffusion} fuses diffusion paths over spatial windows to enable spatially controlled generation. DiffCollage~\cite{zhang2023diffcollage} similarly composes large content through parallel diffusion generation, while SyncDiffusion~\cite{lee2023syncdiffusion} synchronizes overlapping diffusion windows to decrease window incoherence for separate patches. Subsequent panorama and synchronized-generation methods further develop this idea for wide fields of view and semantically coherent compositions~\cite{ye2024diffpano,wang2024customizing,quattrini2024merging,zhang2024taming,kim2024synctweedies}. These works show that increasing canvas size or detail density is not only a matter of producing more pixels: local generation must be coordinated so that fine detail does not fragment the global scene.

Prompt- and strength-based controls are also well established in research and artistic tools. Mixture of Diffusers~\cite{jimenez2023mixture} assigns prompts to spatial regions and harmonizes their diffusion processes across a larger canvas, while ControlNet Tile~\cite{zhang2023controlnettile} and Ultimate SD Upscale~\cite{coyote2023ultimate} combine controlled tiled refinement with denoising-strength controls. In contrast to these sequential refinement approaches, \METHODNAME{} jointly denoises coupled LR and HR trajectories, allowing the evolving representations to interact bidirectionally at every step rather than treating a completed LR image as a fixed input to HR refinement.

A second branch studies tuning-free high-resolution generation with pretrained generative models. DemoFusion~\cite{du2024demofusion} builds on pretrained latent diffusion and combines progressive upscaling, skip residuals, and dilated sampling to produce high-resolution outputs without retraining. ScaleCrafter~\cite{he2024scalecrafter} analyzes direct high-resolution generation with pretrained diffusion models, attributing failures such as repetition and structural artifacts partly to limited effective receptive fields and introducing inference-time changes such as re-dilation. AccDiffusion~\cite{lin2024accdiffusion} targets semantic inconsistency and object repetition in higher-resolution generation. More recent work moves this problem into Diffusion Transformers: ResDiT~\cite{ma2026resdit} studies positional-embedding behavior, local enhancement, patch-level fusion, and layout preservation for resolution extrapolation, while SEGA~\cite{rajabi2026sega} proposes spectral-energy guided attention to balance global structure and fine-detail fidelity in DiT-based extrapolation. They reveal a recurring tension between local detail, global coherence, computational cost, and adherence to conditioning signals, which is precisely the regime in which high-complexity artistic image generation becomes difficult.

\section{Method}
\label{sec:method}

\begin{figure*}[t!]
  \centering
  \resizebox{\linewidth}{!}{\input{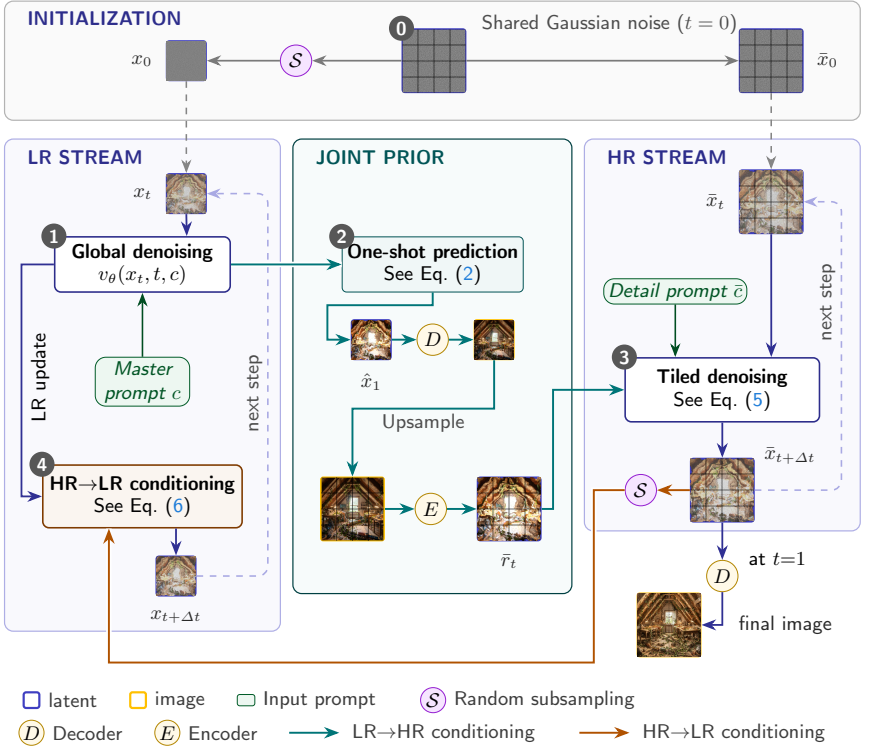}}
  \caption{\textbf{Overview of \METHODNAME.} At $t=0$, a shared Gaussian draw initializes the HR canvas $\bar{x}_0$ and is randomly subsampled to initialize the LR latent $x_0$. Two denoising trajectories then run jointly over the same $4$ steps of a distilled flow-matching model. The LR stream denoises the whole scene from the master prompt; the HR stream denoises overlapping windows and adds local detail. At every step, the LR stream's clean estimate $\hat{x}_1$ is decoded, upsampled, and re-encoded into a reference latent $\bar{r}_t$ whose aligned crops $F_i(\bar{r}_t)$ condition the HR stream. The updated HR canvas is randomly subsampled and blended into the LR update with weight $\alpha$ before the next step.}
  \label{fig:method}
\end{figure*}

\METHODNAME{} couples a global LR denoising trajectory with a tiled HR trajectory, as seen in \cref{fig:method}. Our method uses LR and HR branches: the former maintains coherent scene composition, while the latter synthesizes local detail. Importantly, these two branches do not operate separately. Instead, they exchange information to synchronize high- and low-level image semantics.

\subsection{Preliminaries}
\label{sec:method:preliminaries}

\paragraph{In-context Flow-Matching Sampling.} Modern T2I generators use the flow matching formulation to iteratively generate an image. Let $v_\theta$ be the T2I model and $x_0\sim\mathcal{N}(0,I)\in\mathbb{R}^{C\times H\times W}$ the starting image latent, where $C$, $H$, and $W$ denote its number of channels, height, and width. Let $t$ be the current timestep, $\Delta t$ the timestep size, and $c$ the text prompt. Sampling follows:
\begin{equation}\label{eq:update}
    x_{t + \Delta t} = x_{t} + \Delta t \cdot v_\theta(x_t, t, c),
\end{equation}
until $t=1$.
We can also compute a one-shot estimate of the clean latents with:
\begin{equation}\label{eq:one-shot}
    \hat{x}_1(x_{t}, t, c) = x_{t} + (1 - t)\cdot v_\theta(x_{t}, t, c).
\end{equation}
The formulation in \cref{eq:one-shot} is equivalent to \cref{eq:update} by choosing $\Delta t=1-t$. Let $D$ and $E$ be the decoder and encoder of a variational autoencoder (VAE). When $t=1$ (or $\hat{x}_1$ is created), the image is decoded into pixel space with $D(x_1)$. Optionally, $v_\theta$ can also take a reference image $r$ (encoded with $E$), \ie, $v_\theta$ takes as input $(x_t,t,c)$ or $(x_t,t,c,r)$.

\paragraph{MultiDiffusion.}
MultiDiffusion~\cite{bar2023multidiffusion} synthesizes spatially larger latents than the native denoising field by running the base model on overlapping spatial windows and reconciling their local flow predictions. We refer to this higher-resolution latent as a canvas. Formally, suppose that we have a noised latent canvas $\bar{x}_t\in\mathbb{R}^{C\times \bar{H}\times \bar{W}}$ at timestep $t$, with $\bar{H}>H$ and $\bar{W}>W$ as its height and width. Let $F_i(\cdot)$ be a cropping operation and $F_i^\top(\cdot)$ a zero-padding function, both operating in the $i^\text{th}$ window (out of $N$). MultiDiffusion combines the outputs of all $N$ windows as follows:
\begin{equation}\label{eq:multidiff}
    \bar{v}_\text{MD}(\bar{x}_t, t, c) = \frac{\sum_{i=1}^N F_i^\top\big(  w \odot v_\theta(F_i(\bar{x}_t), t, c)    \big)}{\sum_{i=1}^N F_i^\top(w)},
\end{equation}
where $w\in\mathbb{R}^{H\times W}$ are cosine border-ramp weights that smooth transitions between regions to avoid any visible seams. For the rest of the paper, a bar ($\bar{\cdot}$) denotes an HR variable.

\subsection{Joint LR and HR trajectories}
\label{sec:method:stream}

We propose a two-stream approach: each stream transfers information to the other at every iteration. Adopting the variables from the previous section, let $x_t$ and $\bar{x}_t$ be the LR and HR latents, and let $c$ and $\bar{c}$ be the master and detail prompts. The former describes the overall content of the scene, while the latter describes the details to add. Each iteration begins by computing the LR velocity $v_\theta(x_t,t,c)$ and updating $x_t$ using \cref{eq:update}; the resulting global estimate then conditions the HR branch as described below.

\paragraph{\textcolor{joltLRtoHR}{LR$\boldsymbol{\rightarrow}$HR conditioning.\textsuperscript{*}}}
\begingroup
\renewcommand{\thefootnote}{*}
\makeatletter\def\Hy@footnote@currentHref{Hfootnote.colorlegend}\makeatother
\footnotetext{These colors match the color coding used in the method overview (\cref{fig:method}).}
\endgroup
We produce a one-shot approximation $\hat{x}_1$ using \cref{eq:one-shot}. Using $D$, we decode the LR one-shot latents to pixel space, upsample them, and then encode them back to latent space using $E$, that is
\begin{equation}
    \bar{r}_t = E\Big(\texttt{upsample}\big(D(\hat{x}_1)\big)\Big).
\end{equation}
We call $\bar{r}_t$ our joint prior. We deliberately resize in pixel space because conventional latent-space upsampling cannot simultaneously preserve the marginal distribution~\cite{chang2024warped}.
The joint prior thus captures a coarse representation of the content in $x_t$.

To update the canvas $\bar{x}_t$ with $\bar{r}_t$, we adapt MultiDiffusion's tiled-denoising strategy to leverage in-context T2I conditioning using $\bar{r}_t$. For each window $i$, we crop the reference image, $F_i(\bar{r}_t)$, and forward it alongside the canvas crop $F_i(\bar{x}_t)$. Consequently, we modify \cref{eq:multidiff} to include $\bar{r}_t$ as follows:
\begin{equation}\label{eq:hr-branch}
    \bar{v}_\theta(\bar{x}_t, t, \bar{c}, \bar{r}_t) = \frac{\sum_{i=1}^N F_i^\top\big(  w \odot v_\theta(F_i(\bar{x}_t), t, \bar{c}, F_i(\bar{r}_t))    \big)}{\sum_{i=1}^N F_i^\top(w)}.
\end{equation}
Finally, we update $\bar{x}_t$ with $\bar{v}_\theta(\bar{x}_t,t,\bar{c},\bar{r}_t)$ using the standard update schedule (\cref{eq:update}). This sequence produces a consistent trajectory and prevents divergence among the different windows.

\paragraph{\textcolor{joltHRtoLR}{HR$\boldsymbol{\rightarrow}$LR conditioning.\textsuperscript{*}}}
In the reverse direction, we update the LR branch to reflect the details already generated in the HR branch. Once both branches have taken their step, we randomly subsample the canvas $\bar{x}_{t+\Delta t}$ to preserve its marginal latent distribution and inject it into $x_{t+\Delta t}$. More specifically, we linearly mix both branches and then correct the variance to match that of the LR latents using
\begin{equation}\label{eq:hr-to-lr-conditioning}
\begin{split}
    x^\text{temp} =& \, (1 - \alpha) \, x_{t+\Delta t} + \alpha \, \mathcal{S}(\bar{x}_{t+\Delta t}) \\
    x_{t+\Delta t} \leftarrow& \, \mu(x_{t+\Delta t}) + \frac{\sigma(x_{t+\Delta t})}{\sigma(x^\text{temp})}(x^\text{temp} - \mu (x^\text{temp})),
\end{split}
\end{equation}
where $\mathcal{S}$ is a subsampling mechanism, $\alpha\in [0,1]$ is the HR$\rightarrow$LR conditioning strength, and $\mu$ and $\sigma$ denote per-channel spatial mean and standard deviation, broadcast back to the latent grid. To initialize the two latents $x_0$ and $\bar{x}_0$, we do not sample them independently. Instead, we sample $\bar{x}_0$ and use $\mathcal{S}$ to create $x_0=\mathcal{S}(\bar{x}_0)$. The additional LR trajectory requires one model evaluation for every $N$ HR window evaluations, yielding a relative overhead of $1/N$ that becomes negligible for large images.

\section{Experiments}
\label{sec:experiments}

\subsection{Implementation Details and Baselines}
\label{sec:experiments:setup}

We use a four-step-distilled FLUX.2 Klein-9B backbone~\cite{flux2_2026}. Unless stated otherwise, we generate a $4096^2$ px canvas using $1024^2$ px HR windows with a 512 px stride in both dimensions, \ie, 50\% overlap. The LR stream operates at $1024^2$ px, corresponding to a downsampling factor $s=4$. Every run uses a single constant $\alpha$ throughout the four sampling steps; unless stated otherwise, $\alpha=0.25$.

Our full method recomputes the LR-derived in-context prior at every denoising step (joint-prior conditioning), uses $\alpha=0.25$ for HR$\rightarrow$LR conditioning, and applies MultiDiffusion (MD) aggregation in the HR stream, producing $4096^2$ px images. We choose the most representative approaches from the literature as our baselines: DemoFusion~\cite{du2024demofusion}, SEGA~\cite{rajabi2026sega}, and ResDiT~\cite{ma2026resdit}. For fairness, we re-implement every baseline with the same FLUX.2 Klein-9B four-step backbone, isolating the HR strategy from the base model.

\subsubsection{Evaluation Dataset}
\label{sec:experiments:dataset}

We use GPT-5.6 Sol-high to generate 200 semantically dense evaluation prompts. Each prompt is 30 to 80 words long (36.3 on average) and combines at least four otherwise disparate objects, activities, or settings through explicit spatial, functional, or visual relationships. We release the prompt set in the project repository.

\subsubsection{Metrics}
We assess three complementary axes:

\emph{Complexity} metrics measure the degree of image detail according to our definition. We report IC9600~\cite{feng2023ic9600}, Discrete Cosine Transform energy (DCT)~\cite{ahmed1974discrete}, and JPEG ratio~\cite{wallace1991jpeg}; higher values indicate greater complexity rather than universally better quality. DCT and JPEG ratio are frequency- and compression-based complexity proxies, respectively; additional measures are reported in \cref{sec:appendix:extended-complexity}.

\emph{Text alignment} metrics assess correspondence between the generated image and its prompt. We use LMM4LMM-Correspondence~\cite{wang2025lmm4lmm} (LMM corr.): it is well suited to detailed correspondence judgments on our long prompts, and it is the alignment judge that agrees best with the human prompt-match preferences of our user study (\cref{sec:experiments:human-preference}). Additional alignment metrics and their agreement analysis appear in \cref{sec:appendix:extended-alignment,sec:appendix:metric-agreement}.

Finally, \emph{aesthetics and quality} metrics assess visual appeal, composition, and the absence of artifacts. We report LMM4LMM-Perception (LMM percept.) and the Qwen-Image-Bench (QIB) Quality score~\cite{li2026qwen}, which aggregates its three level-2 dimensions: Realism, Detail, and Resolution. All reported metrics are higher-is-better (or higher-is-more-complex). Metrics excluded from our primary claims because of resolution or context limitations are reported in \cref{sec:appendix:legacy}. We complement these axes with a global--local coherence comparison; full protocol details are reported in \cref{sec:appendix:coherence-diagnostics}.

As an additional evaluation, we conduct a user study, described in \cref{sec:experiments:human-preference}.

\subsection{Quantitative Comparison}
\label{sec:experiments:quantitative}

As shown in \cref{tab:main}, several points are evident. (i) Our full method outperforms all baselines on the three complexity metrics shown, with particularly large gains in DCT and JPEG ratio. (ii) Its LMM4LMM correspondence score remains competitive at $0.367$, only $0.004$ below SEGA, and this difference is not significant after metric-family correction. The detail prompt deliberately introduces plausible content beyond the master prompt, which can be penalized by strict correspondence judgments even when it increases local visual complexity. (iii) \METHODNAME{} has the highest mean QIB Quality and LMM4LMM Perception scores, although neither top-two difference is significant after correction. Thus, its added complexity creates a distinct aesthetic without measured quality degradation. \Cref{sec:appendix:radar} summarizes the metric families.

\begin{table}[H]
\centering
\captionsetup{font=small,skip=3pt}
\caption{\textbf{Quantitative comparison on 200 prompts.} Our full method leads all three displayed complexity metrics. Compared with the strongest baseline, the gains are $5\%$ on IC9600, $50\%$ on DCT energy, and $26\%$ on JPEG ratio. It also has the highest means on LMM4LMM perception and QIB Quality. Its LMM4LMM correspondence is within $0.004$ of the leading baseline. Entries are mean $\pm$ sample standard deviation. Boldface and underlining mark the best and second-best means only when their paired difference is significant after metric-family correction ($p_{\mathrm{adj}}<.05$, two-sided Wilcoxon). Latency reports seconds per image on one H100 80 GB}
\label{tab:main}
\begingroup
\setlength{\tabcolsep}{2.8pt}
\renewcommand{\arraystretch}{1.05}
\resizebox{\linewidth}{!}{%
\begin{tabular}{l*{7}{c}}
\toprule
& \multicolumn{3}{c}{Complexity} & \multicolumn{1}{c}{Text alignment} & \multicolumn{2}{c}{Aesthetics \& quality} & \multicolumn{1}{c}{Latency} \\
\cmidrule(lr){2-4}\cmidrule(lr){5-5}\cmidrule(lr){6-7}\cmidrule(lr){8-8}
Method & IC9600 ($\uparrow$) & DCT ($\uparrow$) & JPEG ($\uparrow$) & LMM corr. ($\uparrow$) & LMM percept. ($\uparrow$) & QIB Quality ($\uparrow$) & Time (s) ($\downarrow$) \\
\midrule
SEGA & $0.615\!\pm\!0.070$ & $483\!\pm\!236$ & $0.096\!\pm\!0.022$ & $0.371\!\pm\!0.038$ & $0.423\!\pm\!0.160$ & $47.2\!\pm\!10.4$ & $75\!\pm\!9$ \\
ResDiT & $\underline{0.714\!\pm\!0.073}$ & $490\!\pm\!206$ & $0.095\!\pm\!0.019$ & $0.360\!\pm\!0.043$ & $0.448\!\pm\!0.152$ & $47.3\!\pm\!10.0$ & $86\!\pm\!22$ \\
DemoFusion & $0.689\!\pm\!0.068$ & $\underline{613\!\pm\!219}$ & $\underline{0.104\!\pm\!0.017}$ & $0.356\!\pm\!0.044$ & $0.411\!\pm\!0.167$ & $50.2\!\pm\!9.2$ & $125\!\pm\!10$ \\
\textbf{Ours} & $\mathbf{0.753\!\pm\!0.055}$ & $\mathbf{920\!\pm\!140}$ & $\mathbf{0.131\!\pm\!0.011}$ & $0.367\!\pm\!0.047$ & $0.462\!\pm\!0.164$ & $50.9\!\pm\!9.4$ & $127\!\pm\!15$ \\
\bottomrule
\end{tabular}
}
\endgroup
\end{table}

To assess whether added detail belongs to the surrounding scene, Qwen3.6-27B~\cite{qwen36_27b} compares each full image and four corresponding crops without seeing the generation prompt or method names. \Cref{tab:coherence-vlm} shows that the judge rates \METHODNAME{} in the same range as DemoFusion and significantly prefers it to SEGA and ResDiT ($p_{\mathrm{Holm}}<.001$).

\begin{table}[H]
\centering
\caption{\textbf{Global--local coherence comparison} on 200 paired prompts per baseline. Order agreement is the percentage of pairs for which the decision is unchanged after reversing the A/B presentation order. Intervals are 95\% percentile-bootstrap CIs and $p_{\mathrm{Holm}}$ corrects exact sign tests. More details are provided in \cref{sec:appendix:coherence-diagnostics}.}
\label{tab:coherence-vlm}
\begingroup
\setlength{\tabcolsep}{3.0pt}
\renewcommand{\arraystretch}{1.08}
\begin{tabular}{lcccc}
\toprule
Baseline & Preference [95\% CI] & Order agree.
& $p_{\mathrm{Holm}}$ & Outcome \\
\midrule
DemoFusion & $.458\;[.398,.520]$ & $79.5\%$ & $.204$ & Same range \\
SEGA & $.748\;[.693,.800]$ & $83.5\%$ & $<.001$ & JoLT preferred \\
ResDiT & $.940\;[.910,.965]$ & $93.0\%$ & $<.001$ & JoLT preferred \\
\bottomrule
\end{tabular}

\endgroup
\end{table}

\subsection{Qualitative Comparison}
\label{sec:experiments:qualitative}

\begin{figure}[H]
  \centering
  \begin{minipage}[t]{.245\textwidth}
    \centering
    \textbf{\vphantom{\METHODNAME{} (ours)}\METHODNAME{} (ours)}
    \par\vspace{0.15em}
    \includegraphics[width=\linewidth]{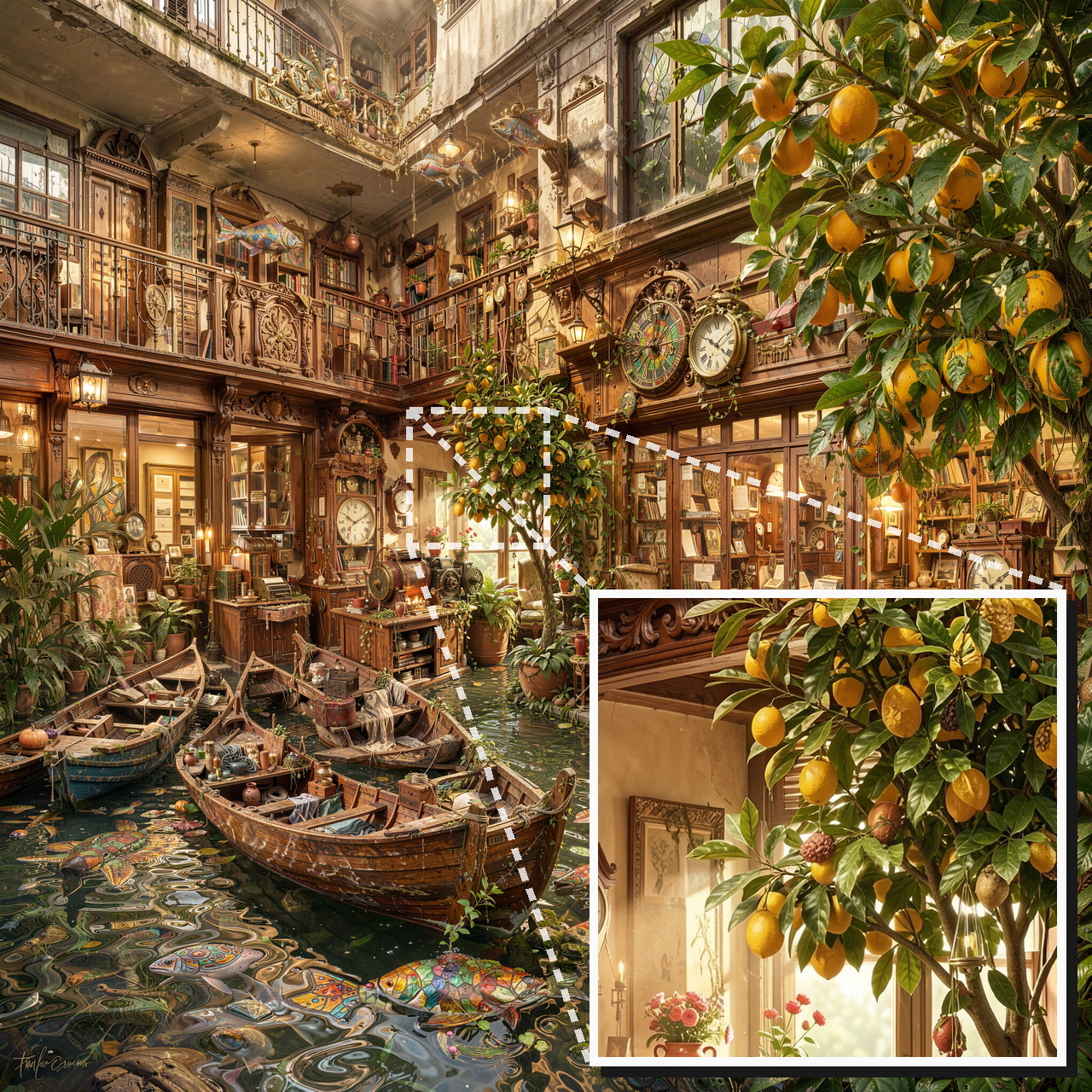}
  \end{minipage}\hfill%
  \begin{minipage}[t]{.245\textwidth}
    \centering
    \textbf{\vphantom{\METHODNAME{} (ours)}DemoFusion}
    \par\vspace{0.15em}
    \includegraphics[width=\linewidth]{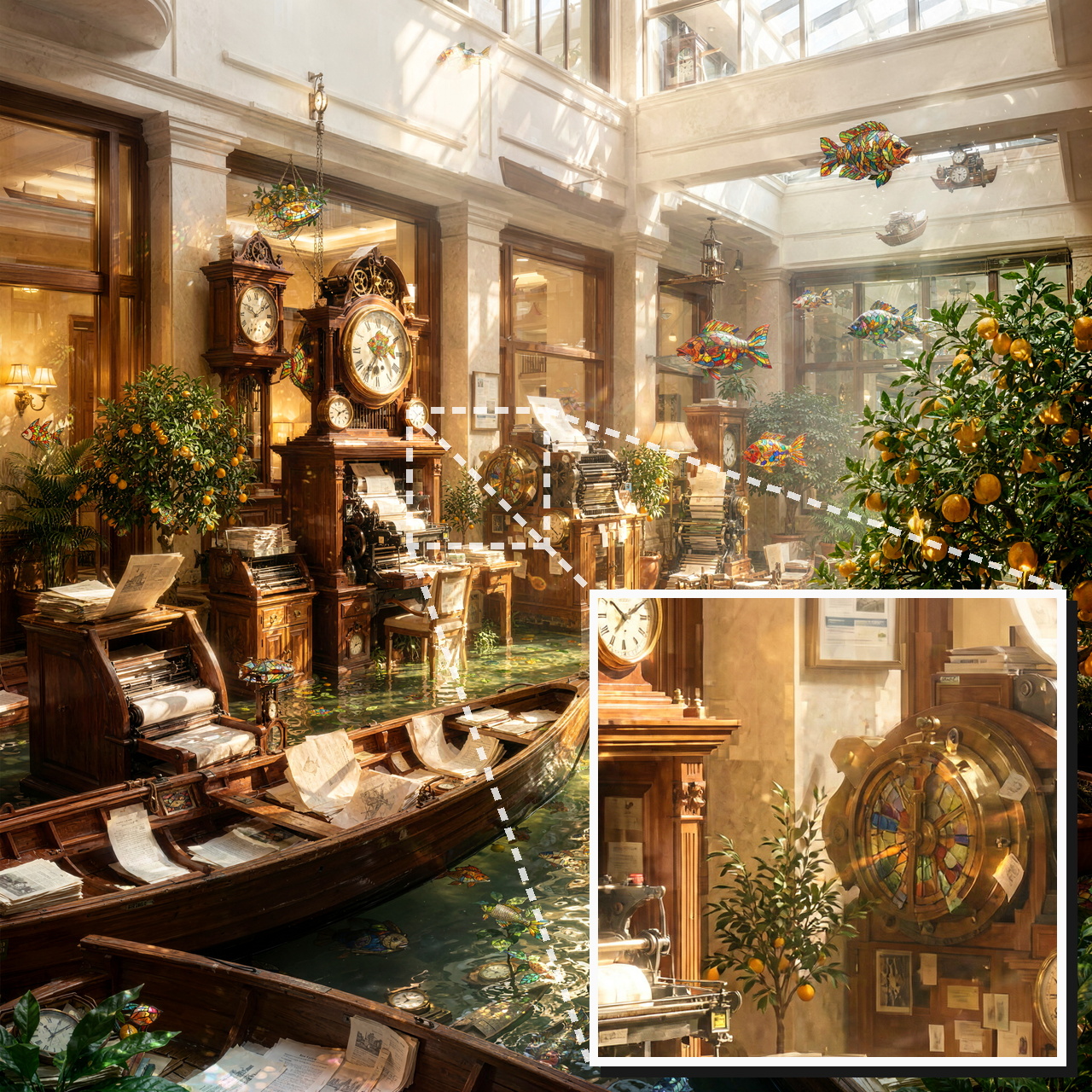}
  \end{minipage}\hfill%
  \begin{minipage}[t]{.245\textwidth}
    \centering
    \textbf{\vphantom{\METHODNAME{} (ours)}ResDiT}
    \par\vspace{0.15em}
    \includegraphics[width=\linewidth]{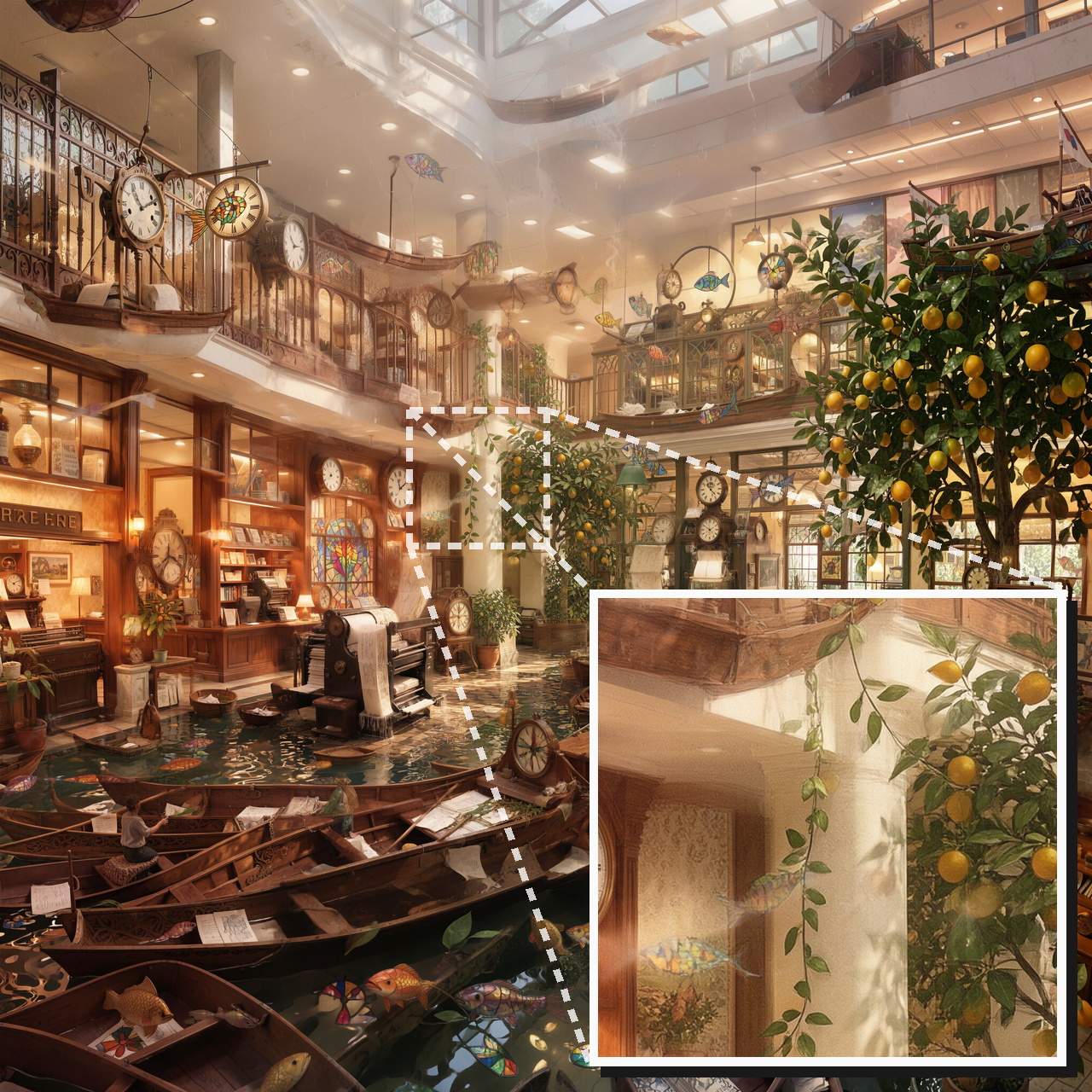}
  \end{minipage}\hfill%
  \begin{minipage}[t]{.245\textwidth}
    \centering
    \textbf{\vphantom{\METHODNAME{} (ours)}SEGA}
    \par\vspace{0.15em}
    \includegraphics[width=\linewidth]{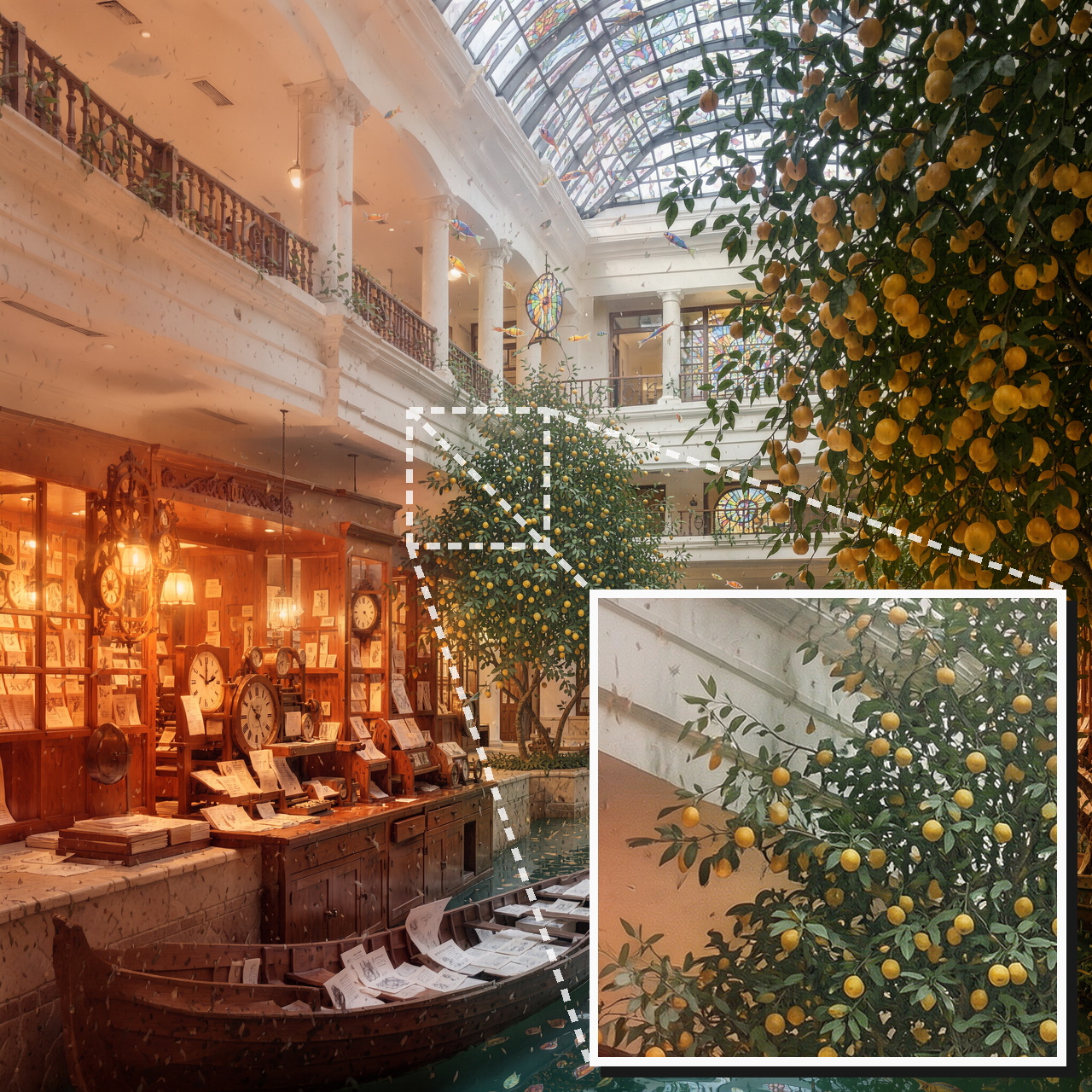}
  \end{minipage}
  \par\vspace{0.3em}
  {\small\itshape A flooded hotel atrium contains a functioning print shop with reed boats, mechanical clocks, stained-glass fish, and citrus trees [...].\par}
  \caption{\textbf{Qualitative comparison with high-resolution generation baselines.} Our result is shown in the leftmost panel. All methods use the same backbone and output resolution. Each lower-right inset magnifies the square center crop spanning one eighth of the source image width and height. Among the compared outputs, ours presents the densest and most varied fine-grained detail in both the full-frame view and the magnified inset.}
  \label{fig:qualitative-comparison}
\end{figure}

Next, we qualitatively compare \METHODNAME{} with the baselines in \cref{fig:qualitative-comparison}; additional examples are provided in \cref{fig:qualitative-comparison-supp}. The full-frame views show global composition, while the center crops highlight local detail. Across these examples, ResDiT tends toward a darker ambience, SEGA often exhibits repetitive textures, and DemoFusion uses simpler layouts. In contrast, \METHODNAME{} produces denser and more varied local structure while retaining a coherent global composition.

Overall, \METHODNAME{} produces striking visuals with substantially greater detail and visual complexity than the compared state-of-the-art methods. Its denser local structure and richer compositions introduce a distinct aesthetic while preserving coherent layouts and strong visual quality.

\subsection{Human Preference}
\label{sec:experiments:human-preference}

\label{sec:experiments:user-study}
\FloatBarrier

Following standard practice in text-to-image generation, we conducted a user study on Amazon Mechanical Turk (AMT) to quantify user preferences against the baselines.\footnote{We restricted participation to AMT workers with the Master qualification and approval rates above 95\%.} For each prompt, a participant evaluated a randomized left/right pair containing our output alongside one of the baselines: DemoFusion, SEGA, or ResDiT. Participants answered three independent questions evaluating which image was (i) more detailed and visually complex, (ii) a better match to the prompt, and (iii) more artistically appealing. In total, the study collected 2,700 pairwise preference judgments across a diverse set of 120 prompts.

\Cref{tab:user-study} summarizes the results, with light-green cells marking significant preferences. (i) Users prefer \METHODNAME{} for detail and complexity against every baseline. (ii) For prompt adherence, \METHODNAME{} is preferred over SEGA and ResDiT, with no significant preference against DemoFusion, consistent with our focus on complexity rather than maximal prompt adherence. (iii) Artistic appeal follows the same pattern. As seen in \cref{fig:qualitative-comparison}, \METHODNAME{} nevertheless produces striking visuals with denser detail and more complex structure than DemoFusion. Together with \cref{tab:main}, these results indicate a distinct aesthetic and added complexity without measured quality degradation, rather than universal aesthetic superiority. Complementary automatic preference results are provided in \cref{sec:appendix:automatic-preference}.

\definecolor{significantgreen}{RGB}{226,239,218}
\begin{table}[H]
\centering
\caption{\textbf{Two-alternative forced-choice user study.} Entries report the percentage of trials preferring our method as the mean $\pm$ the 95\% confidence-interval half-width. A light-green background marks a preference significantly above the 50\% chance level (95\% confidence interval entirely above 50\%).}
\label{tab:user-study}
\small
\setlength{\tabcolsep}{3pt}
\begin{tabular}{lccc}
\toprule
Baseline & \shortstack{More detailed /\\complex (95\% CI)} & \shortstack{Better prompt\\match (95\% CI)} & \shortstack{More artistically\\appealing (95\% CI)} \\
\midrule
SEGA & \cellcolor{significantgreen}$77.3\% \pm 5.1\%$ & \cellcolor{significantgreen}$69.0\% \pm 5.4\%$ & \cellcolor{significantgreen}$68.0\% \pm 5.5\%$ \\
DemoFusion & \cellcolor{significantgreen}$76.3\% \pm 5.1\%$ & $48.5\% \pm 5.6\%$ & $46.3\% \pm 5.7\%$ \\
ResDiT & \cellcolor{significantgreen}$76.0\% \pm 5.1\%$ & \cellcolor{significantgreen}$68.3\% \pm 5.5\%$ & \cellcolor{significantgreen}$71.0\% \pm 5.4\%$ \\
\bottomrule
\end{tabular}%
\end{table}

\subsection{Ablation Studies}
\label{sec:experiments:ablation}

We ablate the contribution of each component. First, we vary the constant HR$\rightarrow$LR conditioning strength $\alpha$ between runs. Next, we compare static and jointly denoised LR-derived priors and isolate the effect of MD aggregation. Finally, we vary the detail prompt $\bar{c}$ to demonstrate controls beyond increasing complexity. The $\alpha$ sweep uses a matched subset of 25 prompts, while the component ablation uses all 200 evaluation prompts.

\subsubsection{HR$\boldsymbol{\rightarrow}$LR Conditioning Strength}

\begin{figure}[!t]
  \centering
  \resizebox{\linewidth}{!}{\input{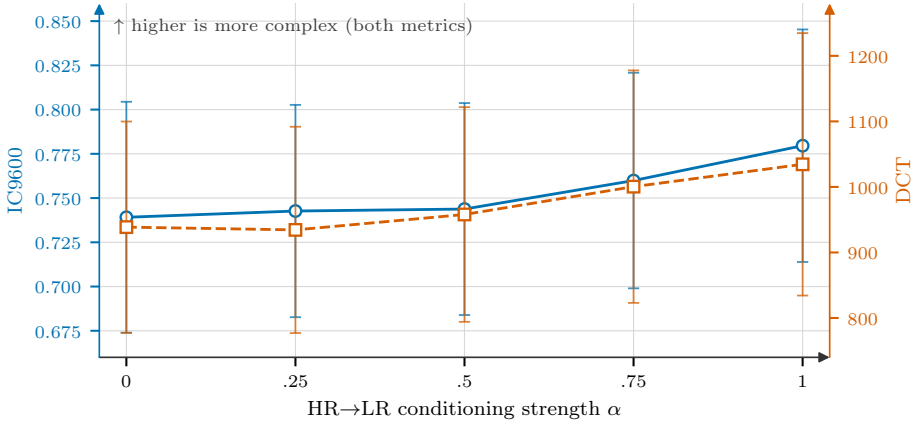}}
  \caption{\textbf{Complexity response to HR$\boldsymbol{\rightarrow}$LR conditioning strength $\alpha$} ($n=25$ matched prompts, tiled denoising at $4096^2$ px final image). Both IC9600 and DCT energy metrics increase most strongly above $\alpha=0.5$, confirming that HR$\rightarrow$LR conditioning provides controllable complexity.}
  \label{fig:alpha-tradeoff}
\end{figure}

\begin{figure*}[!b]
  \centering
  \makebox[\textwidth][l]{%
    \makebox[.2\textwidth][c]{{\bfseries\boldmath $\alpha=0$}}%
    \makebox[.2\textwidth][c]{{\bfseries\boldmath $\alpha=0.25$}}%
    \makebox[.2\textwidth][c]{{\bfseries\boldmath $\alpha=0.50$}}%
    \makebox[.2\textwidth][c]{{\bfseries\boldmath $\alpha=0.75$}}%
    \makebox[.2\textwidth][c]{{\bfseries\boldmath $\alpha=1.0$}}%
  }
  \par\smallskip
  \includegraphics[width=.1975\textwidth]{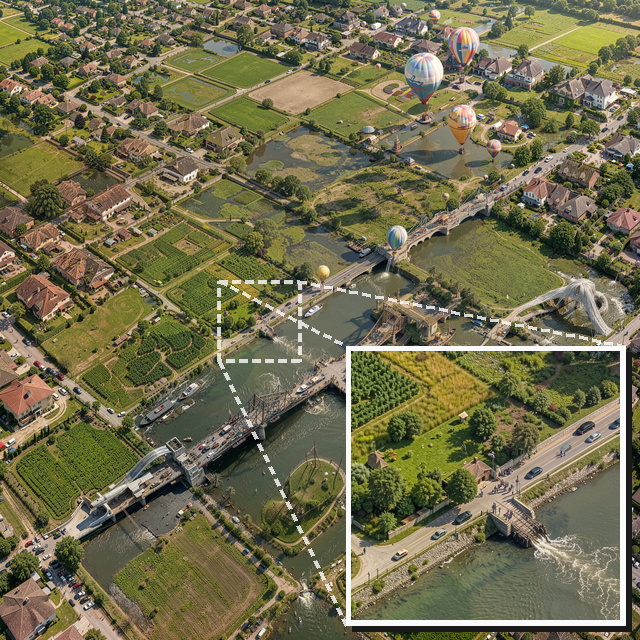}\hfill%
  \includegraphics[width=.1975\textwidth]{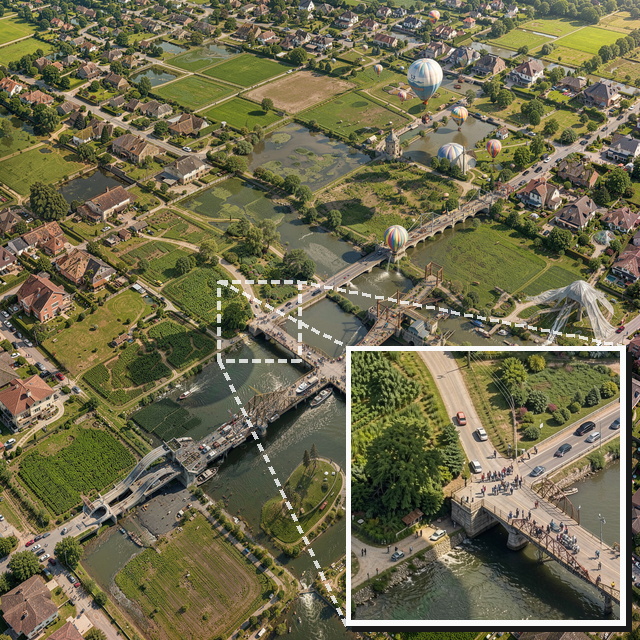}\hfill%
  \includegraphics[width=.1975\textwidth]{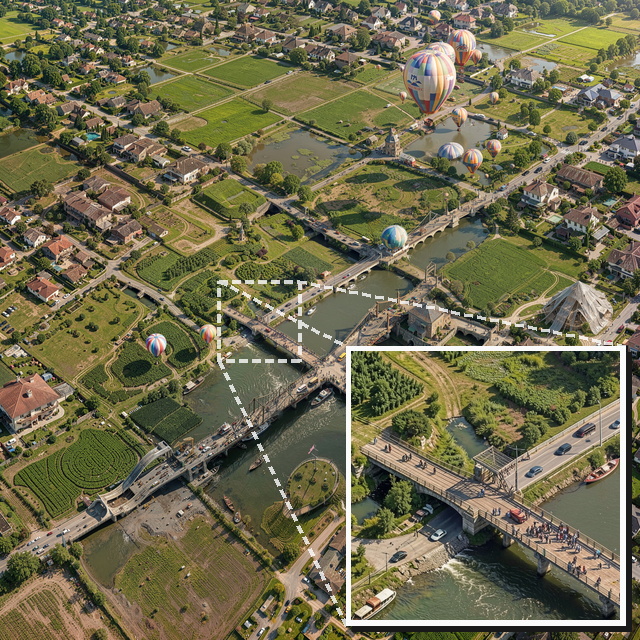}\hfill%
  \includegraphics[width=.1975\textwidth]{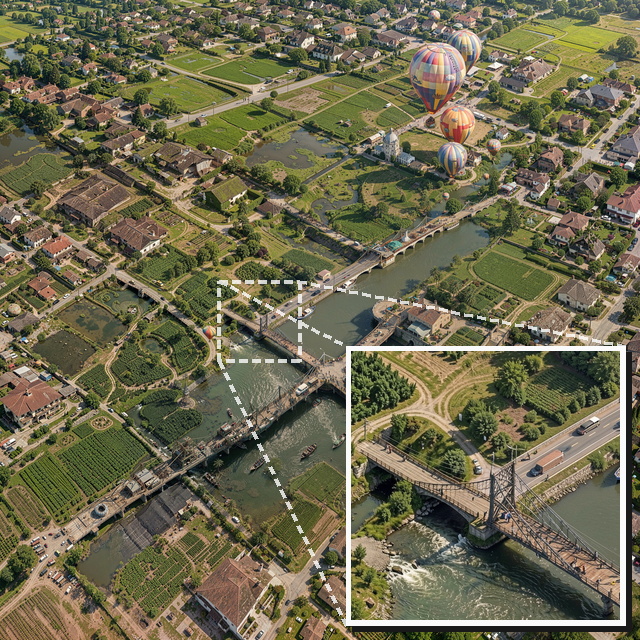}\hfill%
  \includegraphics[width=.1975\textwidth]{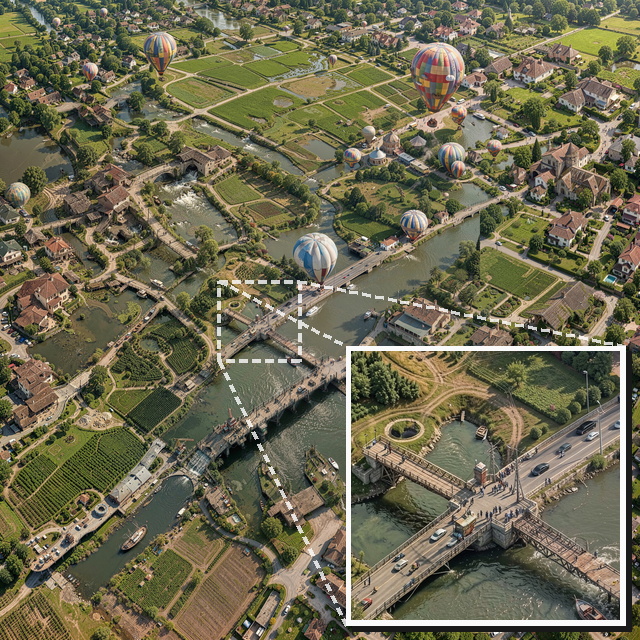}
  \par\smallskip
  {\small\itshape An aerial view reveals a flooded suburb reorganized by violin bridges, rice paddies, meteorological balloons, and concrete playground slides [...].\par}
  \smallskip
  \includegraphics[width=.1975\textwidth]{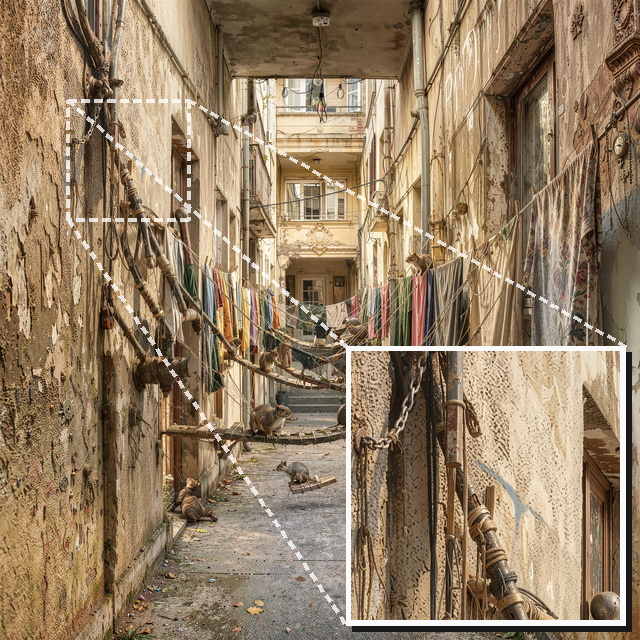}\hfill%
  \includegraphics[width=.1975\textwidth]{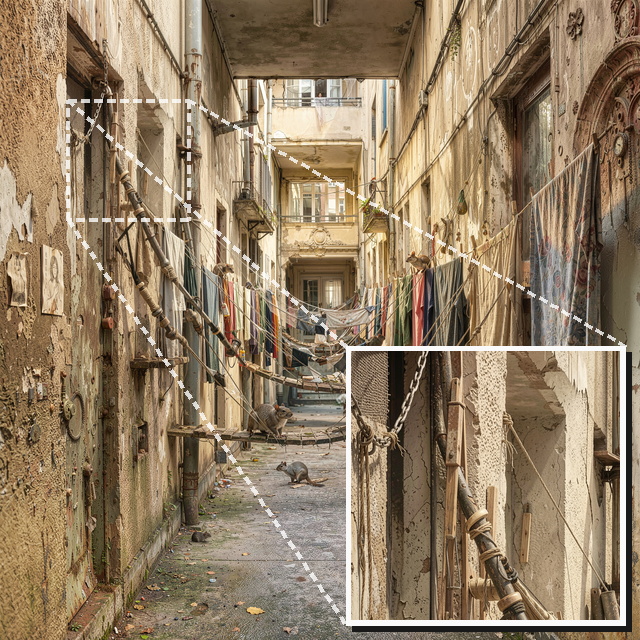}\hfill%
  \includegraphics[width=.1975\textwidth]{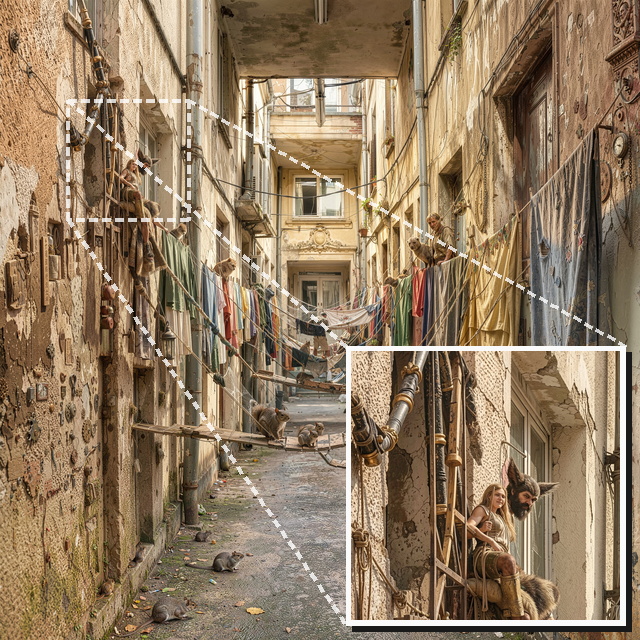}\hfill%
  \includegraphics[width=.1975\textwidth]{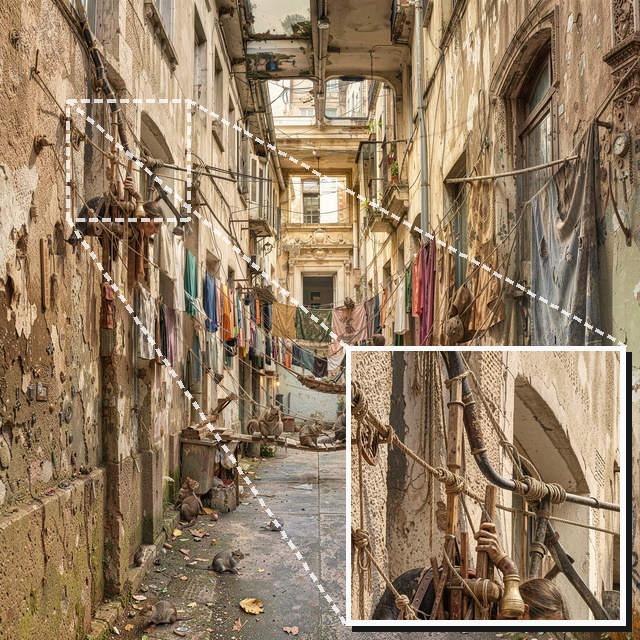}\hfill%
  \includegraphics[width=.1975\textwidth]{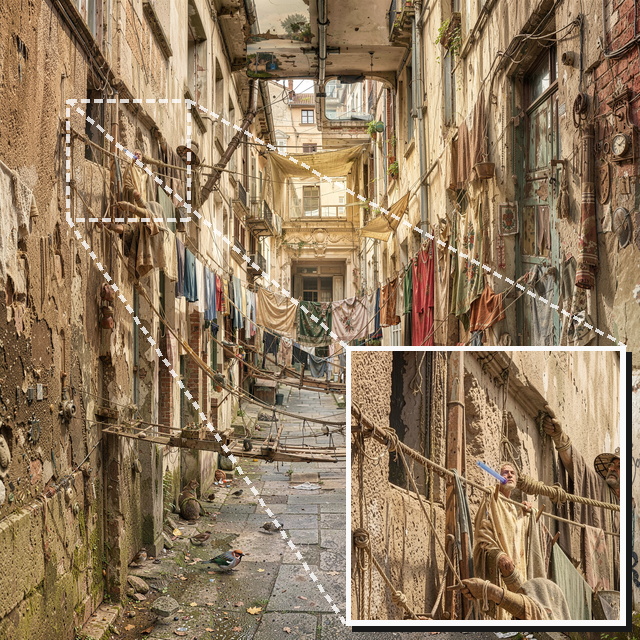}
  \par\smallskip
  {\small\itshape A narrow apartment corridor becomes a choreographed scene for bagpipes, laboratory mice, hanging laundry, and miniature suspension bridges [...].\par}
  \caption{\textbf{Qualitative ablation of HR$\boldsymbol{\rightarrow}$LR conditioning strength $\alpha$.} Across both examples, increasing $\alpha$ produces a clear increase in the number and richness of visible details, especially in the magnified regions shown in the lower-right insets.}
  \label{fig:alpha-qualitative}
\end{figure*}

We evaluate the effect of varying the constant $\alpha$ between runs (\cref{eq:hr-to-lr-conditioning}). When $\alpha=0$, the HR stream still receives information from the LR stream, while the LR stream remains independent of the HR stream. Across the full sweep in \cref{fig:alpha-tradeoff}, IC9600 increases by $5.5\%$ and DCT by $10.2\%$. Both metrics remain near their initial levels through $\alpha=0.5$ before increasing more strongly at $\alpha=0.75$ and $\alpha=1$. We omit the text-alignment curve because LMM4LMM correspondence varies by less than one percentage point across the sweep. We retain $\alpha=0.25$ as a conservative default while leaving headroom for users who prefer more aggressive detail. The matched examples in \cref{fig:alpha-qualitative} likewise show that increasing $\alpha$ produces progressively more complex images, confirming its role as a controllable complexity parameter.

\raggedbottom

\subsubsection{Component Ablation}

We evaluate four configurations to separate joint LR--HR denoising from the effect of MD aggregation. The \emph{prior-only} variant is the base T2I model, without either component. The \emph{static-prior} variant first generates an LR image and then keeps that image fixed as the in-context reference throughout MD sampling. The \emph{joint-prior} variant jointly denoises a $1024^2$ px LR prior and a single $2048^2$ px HR window, updating the LR-derived in-context reference at every step but without MD. Finally, the full \METHODNAME{} combines joint-prior conditioning with MD to produce a $4096^2$ px canvas.

\begin{figure}[!t]
  \centering
  \resizebox{\linewidth}{!}{\input{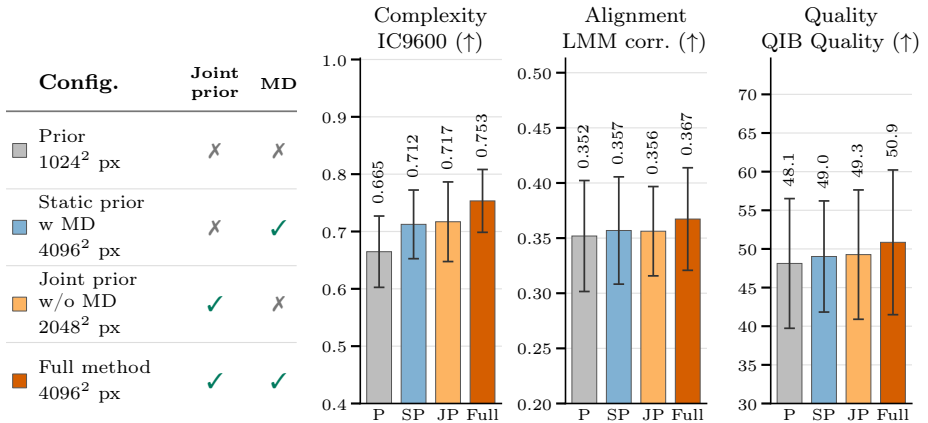}}
  \caption{\textbf{Component ablation.} Left: the four configurations and their native output resolutions; colors match the bars. Right: mean IC9600, LMM4LMM correspondence, and QIB Quality over 200 prompts; error bars show one standard deviation. Since one window is limited to $2048^2$ px, $4096^2$ px generation requires MD.}
  \label{fig:component-ablation}
\end{figure}

These four configurations are summarized in \cref{fig:component-ablation}. Between the two configurations using MD at the same resolution, replacing the static prior with joint-prior conditioning increases IC9600 by $5.7\%$, LMM4LMM correspondence by $2.9\%$, and QIB Quality by $3.7\%$. Relative to joint-prior conditioning in the single-window setting, adding MD increases IC9600 by $5.1\%$, LMM4LMM correspondence by $3.1\%$, and QIB Quality by $3.2\%$. Because one window is limited to $2048^2$ px, evaluation at $4096^2$ px requires MD.

\subsubsection{Detail-Prompt Ablation}

We test the detail prompt as a lightweight editing instruction while holding the master prompt and seed fixed. \Cref{fig:detail-prompt-ablation} shows that the resulting image follows each requested detail or appearance treatment while preserving the same overall scene. This separation lets artists iterate on rendering style, surface texture, and lighting without rewriting the scene description or disrupting its composition. The detail prompt therefore serves as a compact creative-control channel for producing controlled variants of a shared layout.

\begingroup
\setlength{\intextsep}{0pt}
\begin{figure}[H]
  \centering
  \captionsetup{skip=4pt}
  \begin{minipage}[t]{.225\textwidth}
    \centering
    \includegraphics[width=\linewidth]{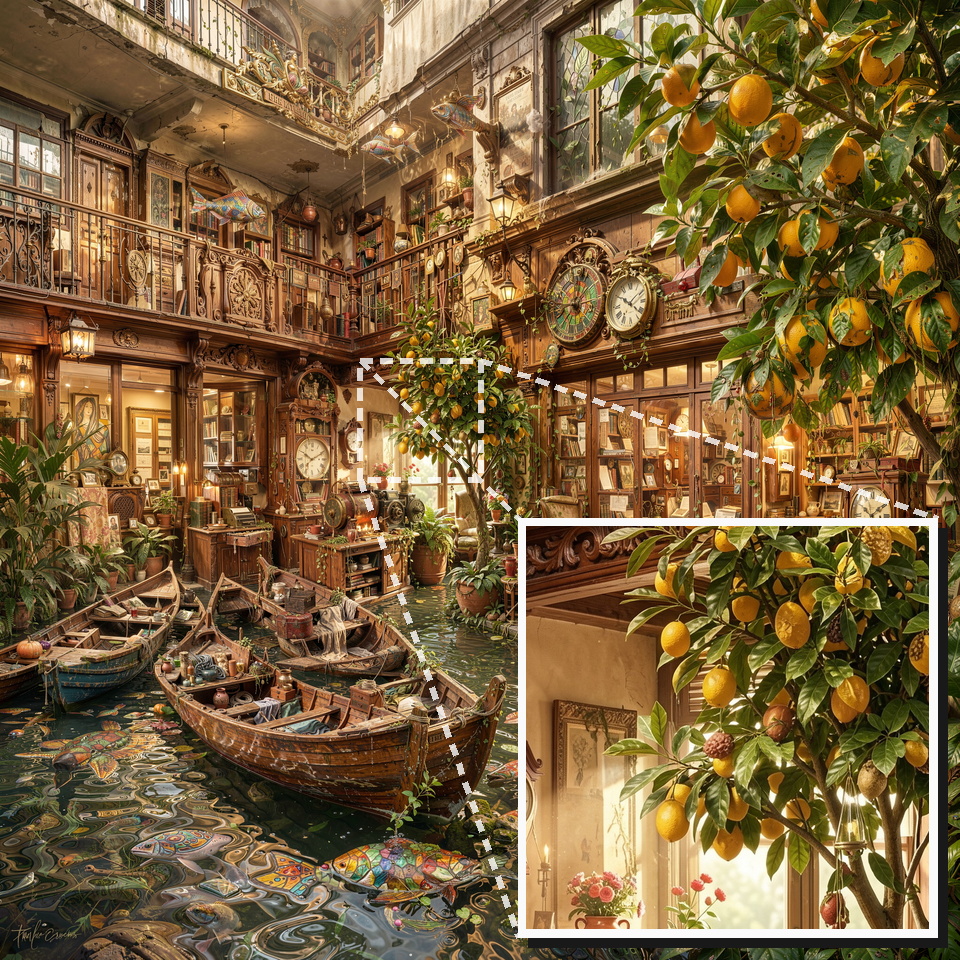}
    \par\vspace{0.25em}
    \parbox[c][3\baselineskip][c]{\linewidth}{\centering\small\textit{``Enhance the image, add lots of details''}}
  \end{minipage}\hfill%
  \begin{minipage}[t]{.225\textwidth}
    \centering
    \includegraphics[width=\linewidth]{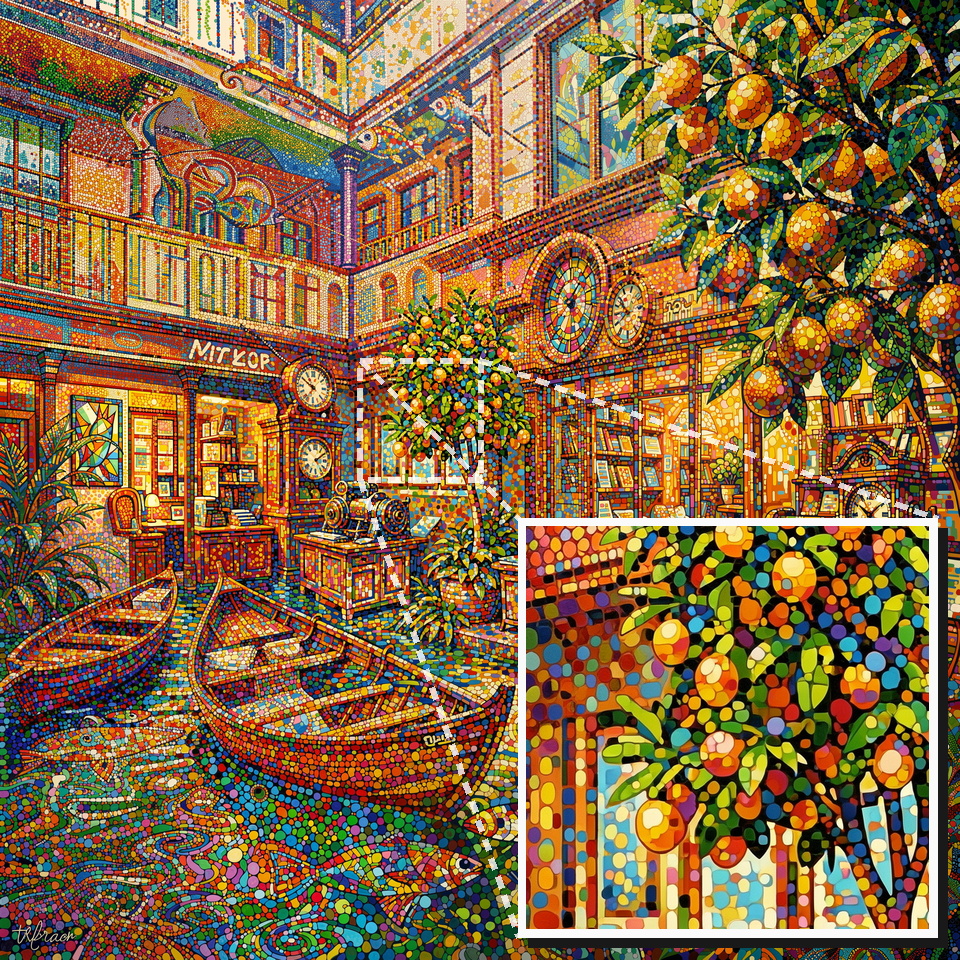}
    \par\vspace{0.25em}
    \parbox[c][3\baselineskip][c]{\linewidth}{\centering\small\textit{``pointillist style''}}
  \end{minipage}\hfill%
  \begin{minipage}[t]{.225\textwidth}
    \centering
    \includegraphics[width=\linewidth]{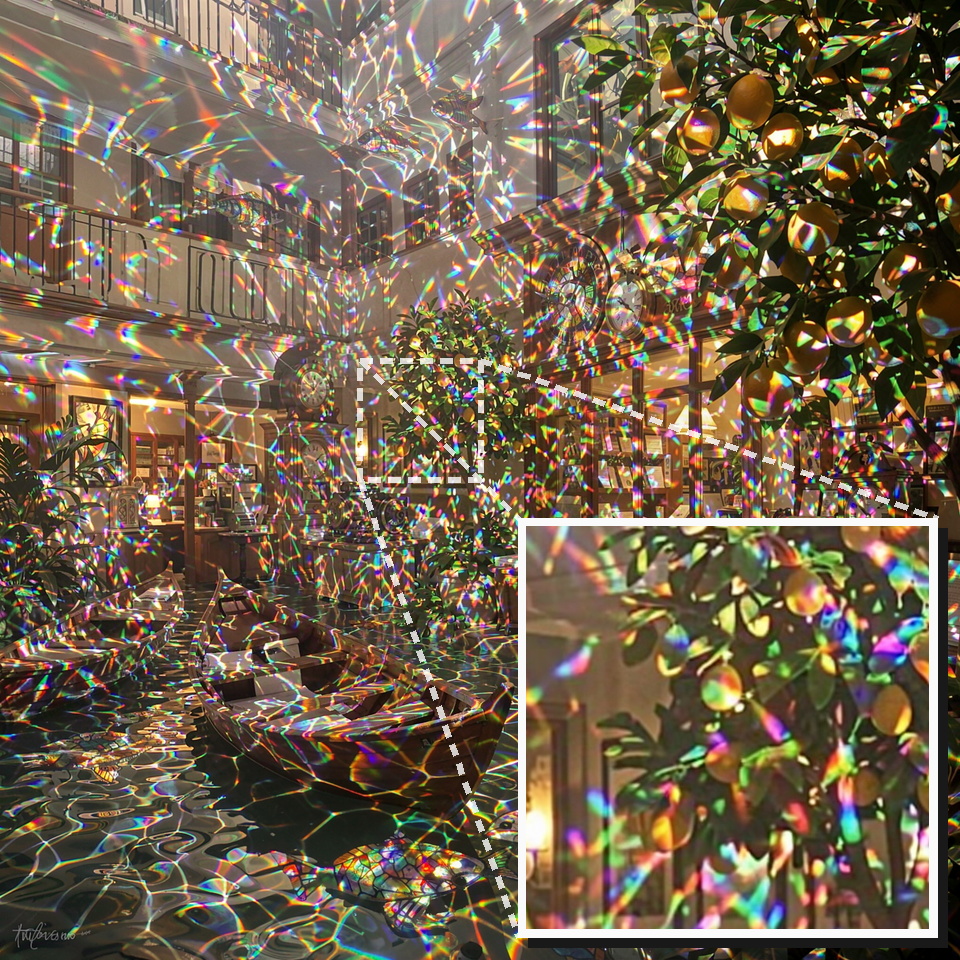}
    \par\vspace{0.25em}
    \parbox[c][3\baselineskip][c]{\linewidth}{\centering\small\textit{``Add iridescent caustic light reflections''}}
  \end{minipage}\hfill%
  \begin{minipage}[t]{.225\textwidth}
    \centering
    \includegraphics[width=\linewidth]{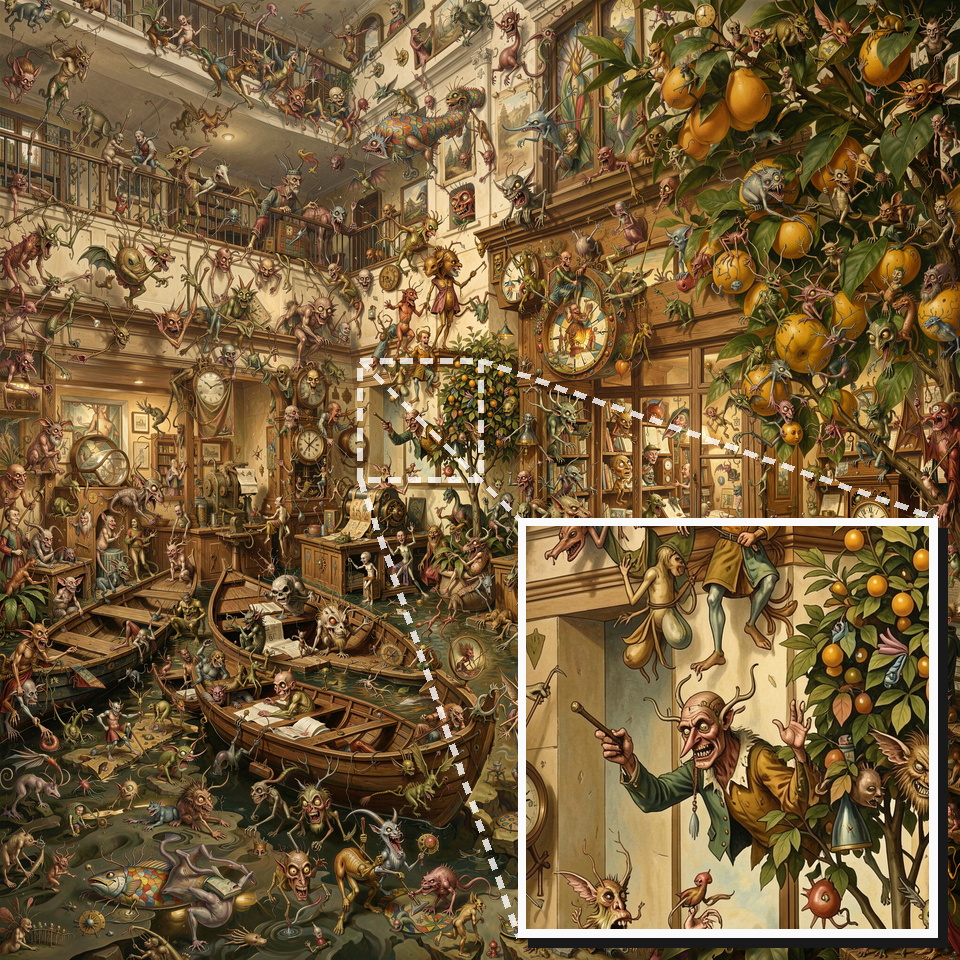}
    \par\vspace{0.25em}
    \parbox[c][3\baselineskip][c]{\linewidth}{\centering\small\textit{``Intricate grotesque imagery''}}
  \end{minipage}
  \caption{\textbf{Qualitative detail-prompt ablation.} We hold the master prompt and seed fixed and vary the detail prompt shown below each image. The master prompt is \emph{``A flooded hotel atrium contains a functioning print shop with reed boats, mechanical clocks, stained-glass fish, and citrus trees [...].''} All results are generated at $4096^2$ px.}
  \label{fig:detail-prompt-ablation}
\end{figure}
\endgroup

\clearpage
\flushbottom

\FloatBarrier
\section{Limitations and Conclusion}
\label{sec:limitations_and_conclusion}

\paragraph{\quad Limitations.}
\label{par:limitations}
Our method relies on the potentially noisy one-shot estimate $\hat{x}_1$. We therefore evaluate it on a four-step model, whose estimate is already visually informative after one step. We also tested our method with longer sampling schedules, but these require an alternate clean-estimation approach, such as PiD~\cite{pid2026}, as well as substantially more compute. Our evaluation is also limited by the weak-to-moderate agreement among text-alignment metrics: across the nine metrics in \cref{sec:appendix:metric-agreement}, their mean pairwise Spearman correlation is only $\rho\approx0.25$, so no single score is a definitive proxy for text--image alignment and metric-specific gains should be read cautiously, ideally corroborated across metrics or with human judgment. We further hypothesize that the sheer density of detail \METHODNAME{} introduces can itself reduce the readability of the full scene: beyond a certain point, additional local content crowds the global composition and makes the whole image harder to parse. We believe this effect partly underlies the modest text-alignment gap we observe alongside our large complexity gains. Future work should also evaluate on larger standardized benchmarks, such as the graph-based DPG-Bench~\cite{hu2024ella} and art-focused GenAI-Bench~\cite{li2024genaibench}.

\paragraph{\quad Conclusion.}
\label{par:conclusion}
We introduced \METHODNAME{}, a two-stream generator that jointly denoises an LR layout trajectory and an HR detail trajectory, coupling them at every step instead of cascading from LR to HR. A global master prompt governs the overall composition while a detail prompt and a step-wise in-context prior drive local synthesis, and a single feedback strength $\alpha$ turns image complexity into a directly controllable parameter. On a distilled four-step FLUX.2 backbone with MultiDiffusion aggregation, \METHODNAME{} generates images at $4096^2$ px and above with substantially higher measured complexity than strong high-resolution baselines while remaining competitive in text alignment, aesthetics, and quality, and human evaluators consistently prefer it for visual detail. More broadly, our results suggest that treating high-resolution synthesis as two coupled trajectories, rather than a low-to-high cascade, is an effective route to dense, high-fidelity imagery. For artistic practice, this turns choices that vanilla T2I models fix implicitly into explicit controls: $\alpha$ acts as a dial over visual density, and the detail prompt restyles a fixed composition through language alone. At resolutions that withstand the close inspection a print invites, \METHODNAME{} becomes a practical instrument for large-format artistic creation.

\makeatletter
\ifeccv@review\else
  \section*{Acknowledgements}

\begin{sloppypar}
  This work has been partially supported by the VISA DEEP chair (grant $\text{ANR-20-CHIA-0022}$), the PostGenAI@Paris cluster (grant $\text{ANR-23-IACL-0007}$; France 2030), and funded by the French National Research Agency (ANR) under the Renaissance project (grant $\text{ANR-23-CE23-0023}$; France 2030).
\end{sloppypar}

This project was provided with computing HPC \& AI and storage resources by GENCI at IDRIS thanks to the grant $\text{2025-AD011017131}$ on the supercomputer Jean Zay’s H100 partition.

A CC BY 4.0 public copyright license has been applied by the authors to the present document and will be applied to all subsequent versions up to the Author Accepted Manuscript (AAM) arising from this submission, in accordance with the grant’s open access conditions.

\fi
\makeatother

\bibliographystyle{splncs04}
\bibliography{main}

\clearpage

\appendix
\makeatletter
\renewcommand{\theHsection}{appendix.\thesection}
\renewcommand{\theHsubsection}{appendix.\thesubsection}
\renewcommand{\theHsubsubsection}{appendix.\thesubsubsection}
\makeatother

\begin{center}
  {\noindent \Large \textbf{JoLT: Joint Latent Trajectories for Context-Guided High-Resolution Tiled Generation}}
\end{center}

\section{Additional Qualitative Comparisons}
\label{sec:appendix:qualitative}

We provide six additional matched comparisons using the same ordering and center-crop protocol as \cref{fig:qualitative-comparison}. All methods receive the same prompt and seed at $4096^2$ px with the FLUX.2 Klein-9B backbone.

\begingroup
\setlength{\LFcapwidth}{\textwidth}
\newlength{\qualitativeCaptionSeparation}
\setlength{\qualitativeCaptionSeparation}{11.117pt}
\newcommand{\qualitativeTile}[1]{%
  \makebox[.25\textwidth][c]{\includegraphics[width=.21\textwidth]{#1}}%
}
\begin{longfigure}{@{}c@{}}
  \makebox[\textwidth][l]{%
    \makebox[.25\textwidth][c]{\textbf{\METHODNAME{} (ours)}}%
    \makebox[.25\textwidth][c]{\textbf{DemoFusion}}%
    \makebox[.25\textwidth][c]{\textbf{ResDiT}}%
    \makebox[.25\textwidth][c]{\textbf{SEGA}}%
  }\\
  \endLFfirsthead
  \makebox[\textwidth][l]{%
    \makebox[.25\textwidth][c]{\textbf{\METHODNAME{} (ours)}}%
    \makebox[.25\textwidth][c]{\textbf{DemoFusion}}%
    \makebox[.25\textwidth][c]{\textbf{ResDiT}}%
    \makebox[.25\textwidth][c]{\textbf{SEGA}}%
  }\\
  \endLFhead
  \makebox[\textwidth][r]{%
    \raisebox{-\footskip}[0pt][0pt]{\small\itshape Additional qualitative comparisons continue on the next page.}%
  }\\
  \endLFfoot
  \endLFlastfoot
  \parbox{\textwidth}{\centering
    \qualitativeTile{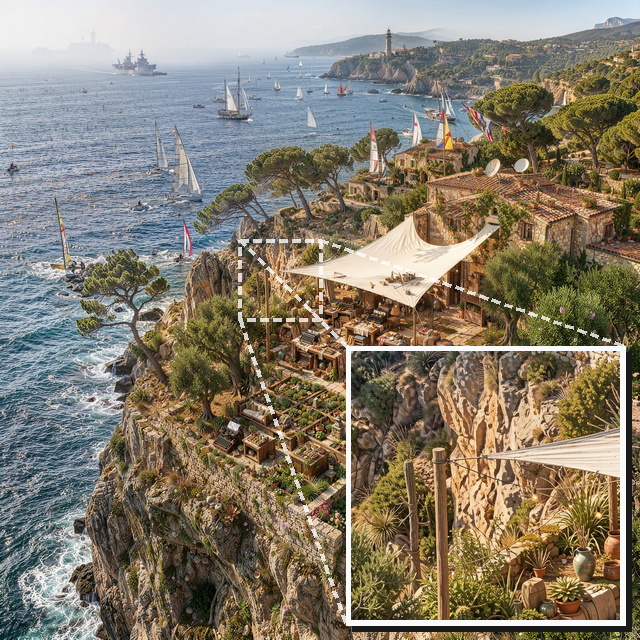}%
    \qualitativeTile{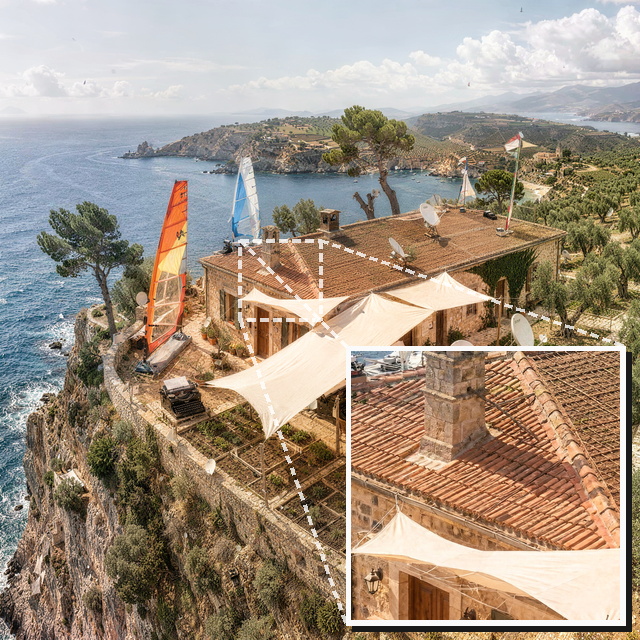}%
    \qualitativeTile{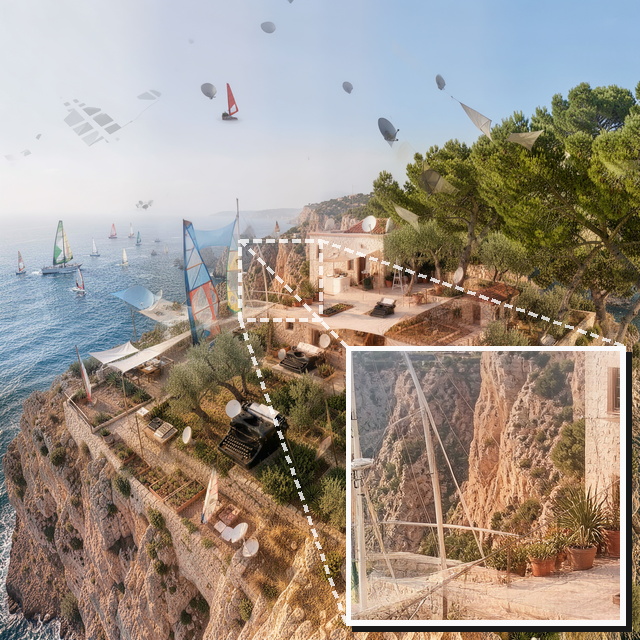}%
    \qualitativeTile{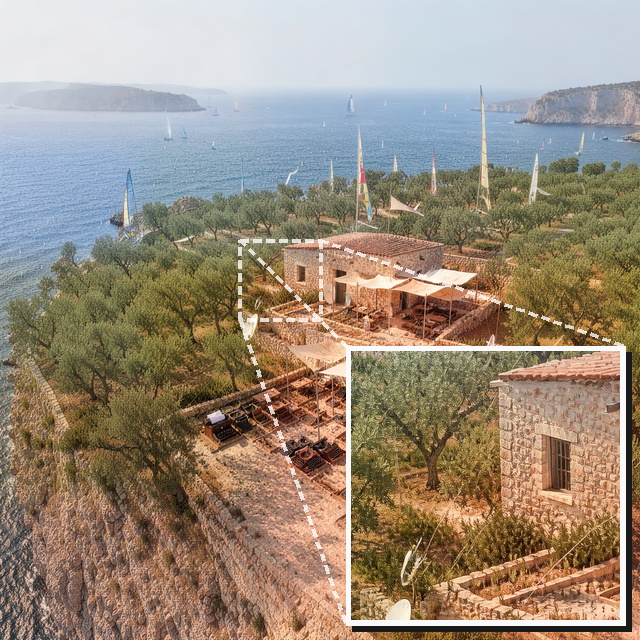}
    \par\smallskip
    {\scriptsize\itshape Viewed from a cliff, a coastal monastery combines windsurfing sails, antique typewriters, olive groves, and satellite dishes [...].\par}
    \smallskip
  }\\
  \parbox{\textwidth}{\centering
    \qualitativeTile{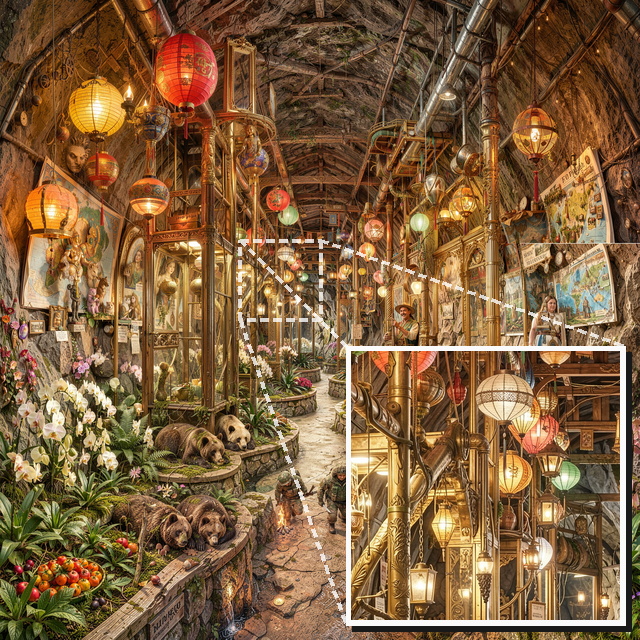}%
    \qualitativeTile{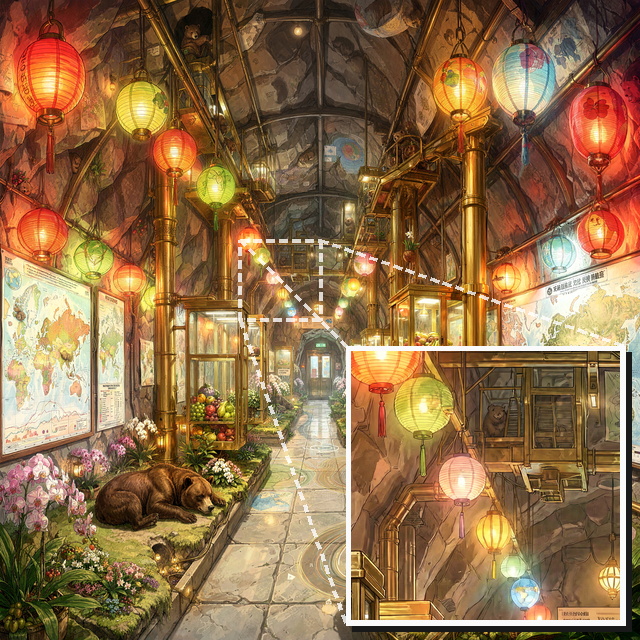}%
    \qualitativeTile{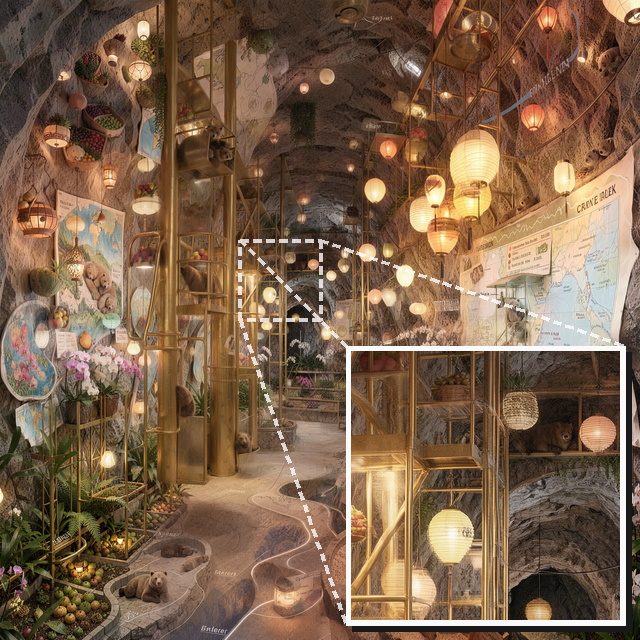}%
    \qualitativeTile{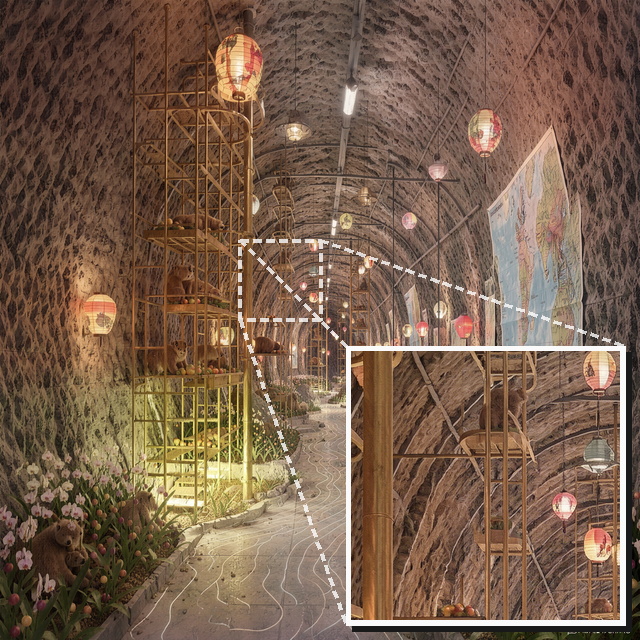}
    \par\smallskip
    {\scriptsize\itshape A mountain tunnel becomes an indoor orchard with brass elevators, paper lanterns, weather maps, and sleeping bears [...].\par}
    \smallskip
  }\\
  \parbox{\textwidth}{\centering
    \qualitativeTile{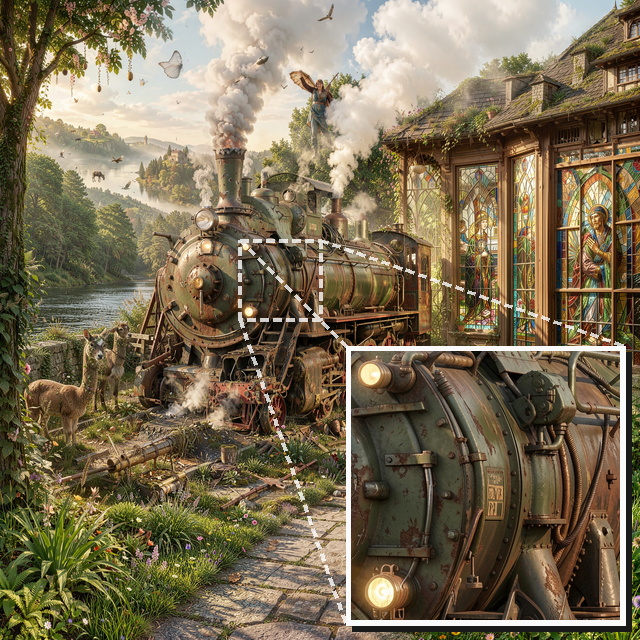}%
    \qualitativeTile{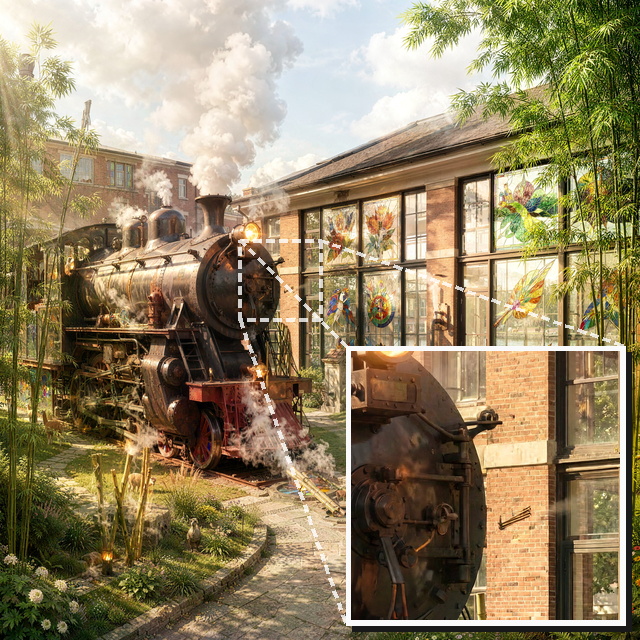}%
    \qualitativeTile{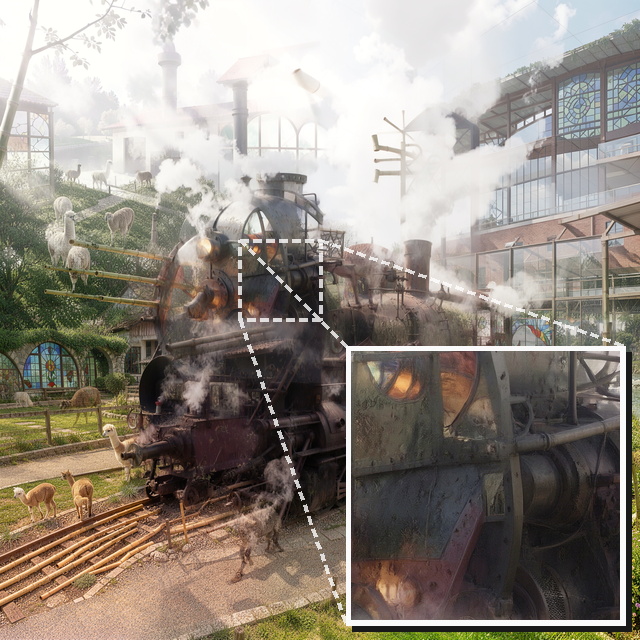}%
    \qualitativeTile{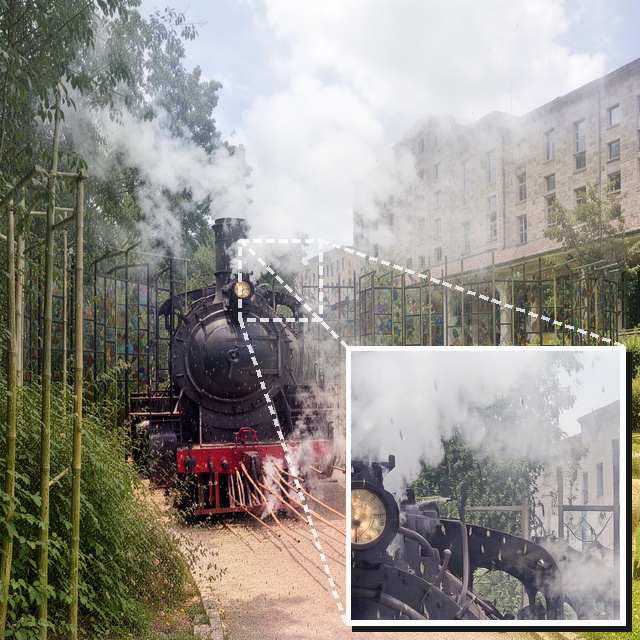}
    \par\smallskip
    {\scriptsize\itshape A riverside hospital garden contains a locomotive boiler, stained-glass screens, bamboo flutes, and grazing alpacas [...].\par}
    \smallskip
  }\\
  \parbox{\textwidth}{\centering
    \qualitativeTile{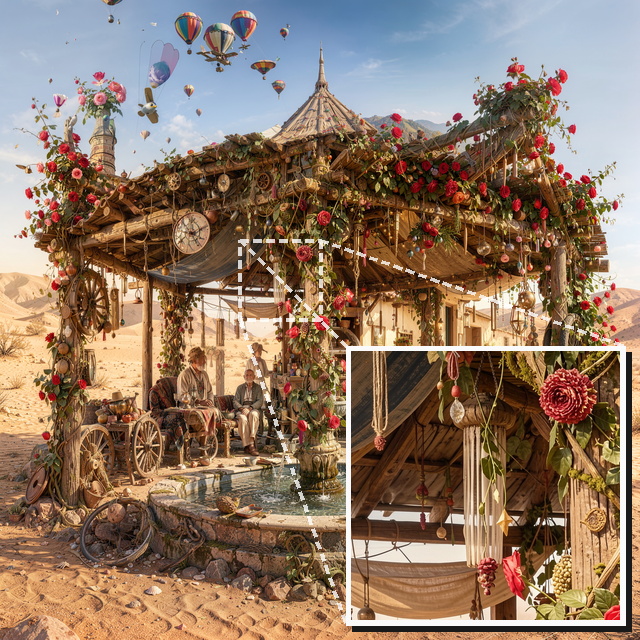}%
    \qualitativeTile{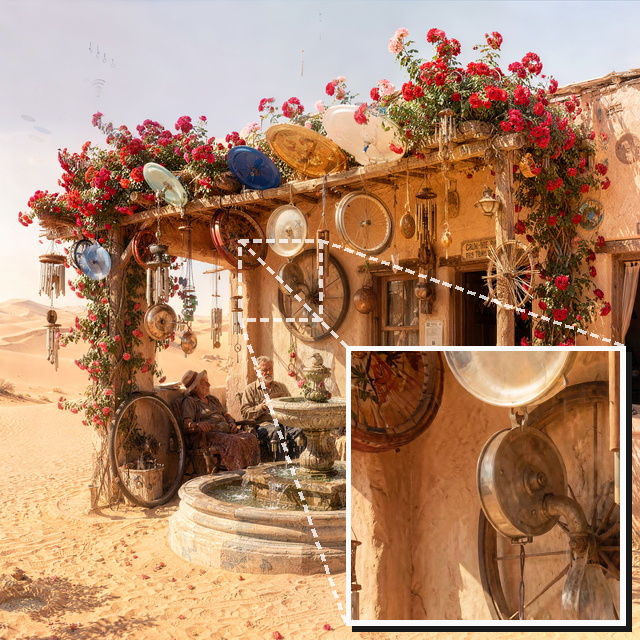}%
    \qualitativeTile{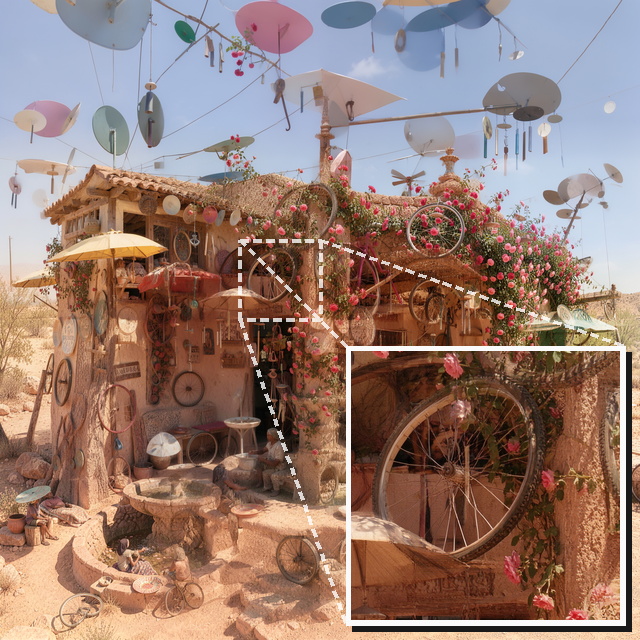}%
    \qualitativeTile{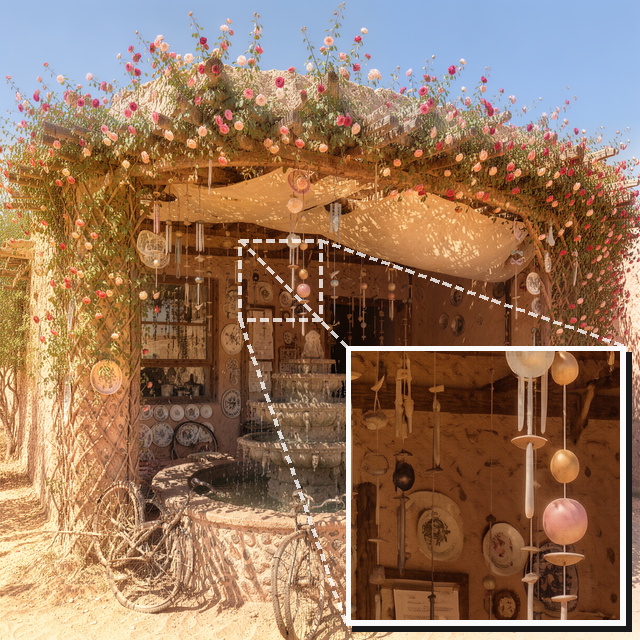}
    \par\smallskip
    {\scriptsize\itshape A desert clinic is built around a ceremonial fountain with x-ray plates, wind chimes, climbing roses, and bicycle wheels [...].\par}
  }\\
  \pagebreak
  \parbox{\textwidth}{\centering
    \qualitativeTile{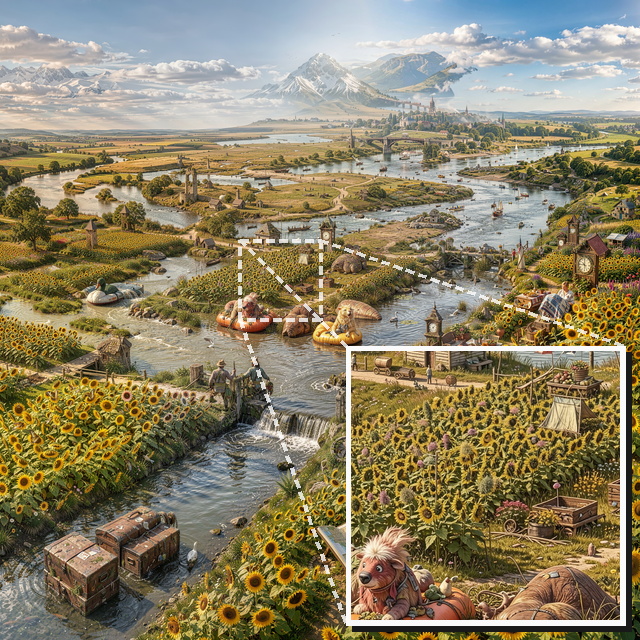}%
    \qualitativeTile{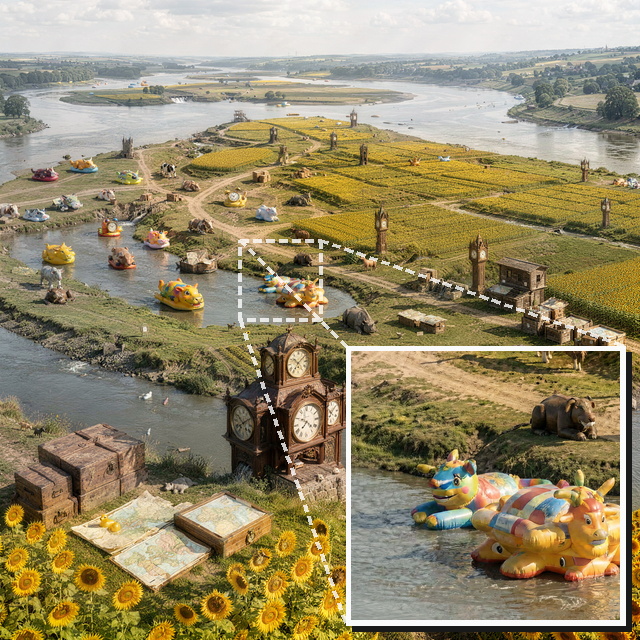}%
    \qualitativeTile{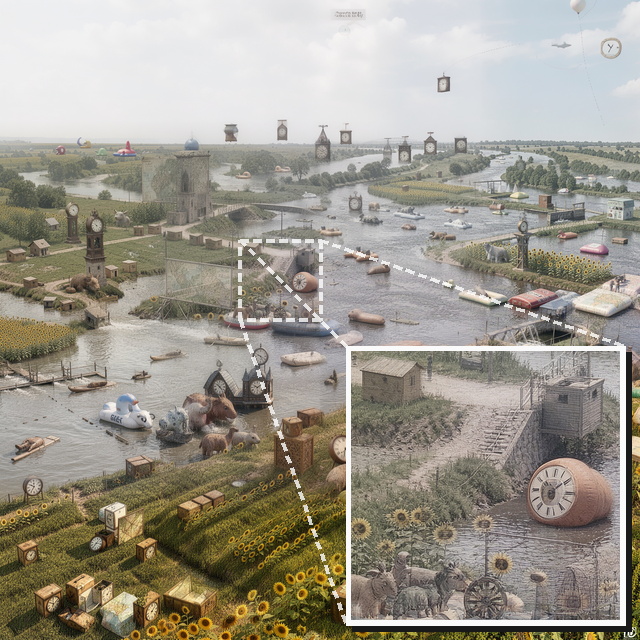}%
    \qualitativeTile{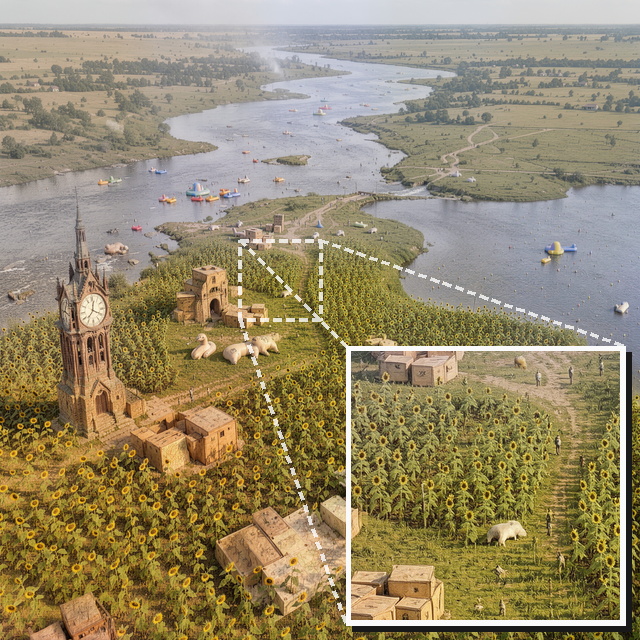}
    \par\smallskip
    {\scriptsize\itshape A broad river delta is surveyed using cathedral clocks, inflatable animals, archival boxes, and sunflower fields [...].\par}
    \smallskip
  }\\
  \parbox{\textwidth}{\centering
    \qualitativeTile{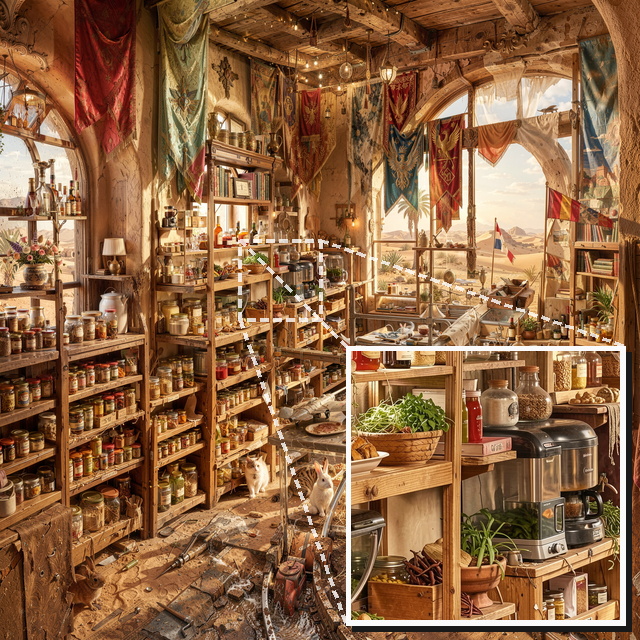}%
    \qualitativeTile{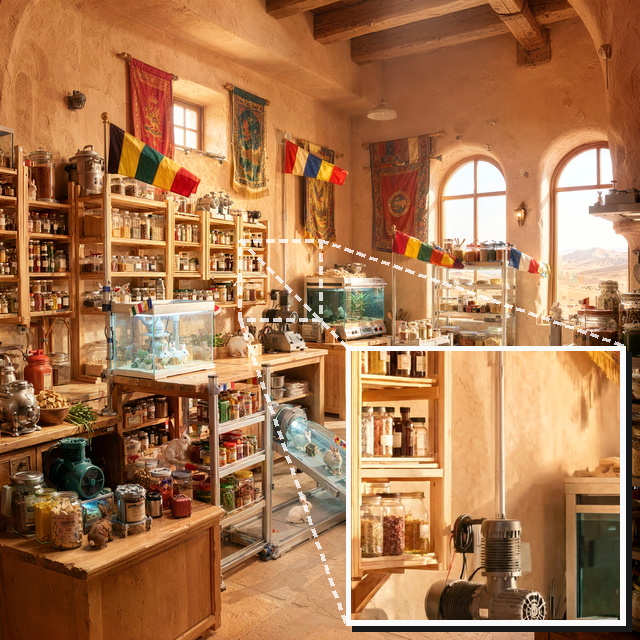}%
    \qualitativeTile{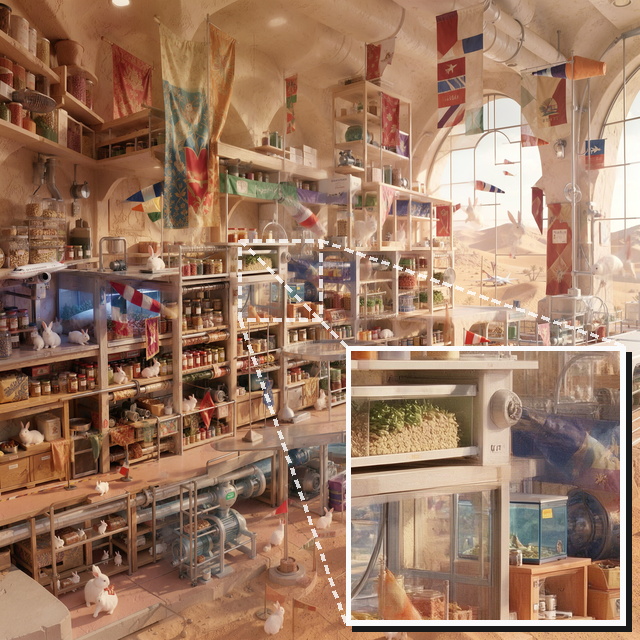}%
    \qualitativeTile{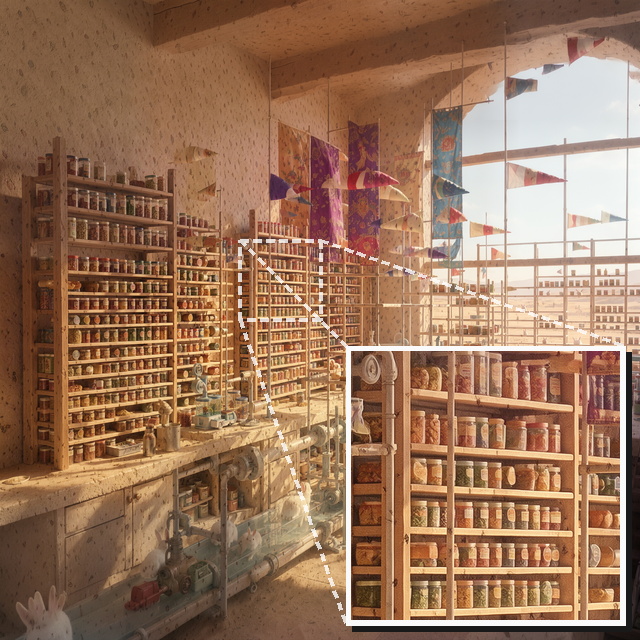}
    \par\smallskip
    {\scriptsize\itshape A desert palace kitchen contains a miniature airport assembled from spice racks, silk banners, aquarium pumps, and white rabbits [...].\par}
    \addvspace{\qualitativeCaptionSeparation}
  }\\
  \caption{\textbf{Additional qualitative comparisons.} Six matched examples and crops at the same position are shown from top to bottom. Our result is leftmost and showcases greater complexity and richer detail.}
  \label{fig:qualitative-comparison-supp}\\
\end{longfigure}
\endgroup

\section{Automatic Preference Evaluation}
\label{sec:appendix:automatic-preference}

We report HPSv2.1~\cite{wu2023humanpreference}, ImageReward~\cite{xu2023imagereward}, and PickScore~\cite{kirstain2023pickapic} as automatic human-preference proxies. All three are poorly suited to our high-resolution setting: each downsamples its input to $224^2$ px before scoring (see \cref{sec:appendix:legacy}), and their scores do not consistently align with our user study (\cref{sec:experiments:user-study}). The three proxies also disagree on the leading method: HPSv2.1 and PickScore place DemoFusion first with our method second, whereas ImageReward places our method first. We therefore include them in \cref{tab:preference} for reference only, not as primary evidence.

\begin{table}[H]
\centering
\caption{\textbf{Automatic human-preference evaluation.} Entries are mean $\pm$ sample standard deviation. Boldface and underlining mark the best and second-best means only when their paired difference is significant after metric-family correction ($p_{\mathrm{adj}}<.05$, two-sided Wilcoxon). Overall, our method performs broadly on par with DemoFusion across the three preference proxies.}
\label{tab:preference}
\small
\setlength{\tabcolsep}{5.0pt}
\begin{tabular}{lccc}
\toprule
Method & HPSv2.1 ($\uparrow$) & ImageReward ($\uparrow$) & PickScore ($\uparrow$) \\
\midrule
SEGA & $0.253\!\pm\!0.030$ & $-0.335\!\pm\!0.246$ & $21.49\!\pm\!0.76$ \\
ResDiT & $0.261\!\pm\!0.027$ & $-0.305\!\pm\!0.233$ & $21.41\!\pm\!0.75$ \\
DemoFusion & $\mathbf{0.304\!\pm\!0.027}$ & $\underline{-0.125\!\pm\!0.228}$ & $\mathbf{22.05\!\pm\!0.74}$ \\
\textbf{Ours} & $\underline{0.293\!\pm\!0.028}$ & $\mathbf{-0.097\!\pm\!0.214}$ & $\underline{21.80\!\pm\!0.77}$ \\
\bottomrule
\end{tabular}

\end{table}

\clearpage
\section{Extended Metric Results and Reliability}
\label{sec:appendix:diagnostics}

This section reports the full metric pool behind the three automatic evaluation axes of \cref{tab:main} and examines how far these metrics can be trusted in our high-resolution, long-prompt setting. \Cref{sec:appendix:coherence-diagnostics} details the direct coherence protocol reported in \cref{tab:coherence-vlm}; \cref{sec:appendix:extended-complexity,sec:appendix:extended-alignment} extend the complexity and text-alignment results of \cref{sec:experiments:quantitative}; \cref{sec:appendix:metric-agreement} quantifies how weakly the text-alignment judges agree with one another; \cref{sec:appendix:radar} summarizes the automatic metric families at a glance; and \cref{sec:appendix:legacy} records legacy metrics whose resolution or context limitations exclude them from the paper's claims. Where applicable, the following multimodal judges use the Qwen3.6-27B backbone~\cite{qwen36_27b}: DSG~\cite{cho2024davidsonian}, VQAScore~\cite{lin2024evaluating}, TIT-Score, and TIT-Score-LLM~\cite{wang2025tit}.

\subsection{Coherence Evaluation}
\label{sec:appendix:coherence-diagnostics}

\paragraph{Crop comparison.}
Pairwise multimodal comparison of text-to-image outputs has precedent in MJ-Bench~\cite{chen2024mjbench}, and prior work shows that such comparisons are sensitive to image order~\cite{hwang2025fooling}. Our crop comparison results are reported in \cref{tab:coherence-vlm}. For each prompt, JoLT is compared separately with each baseline using Qwen3.6-27B~\cite{qwen36_27b}, while the generation prompt and method names are withheld. Each candidate is shown through one $1024^2$ px full view and four $768^2$ px crops: the center and three deterministic 512 px lattice intersections at least 1024 px apart. We ask: \emph{``Which candidate's local content better belongs in and is consistent with the surrounding scene?''}.

To control order effects, each pair is evaluated in randomized A/B order and in reverse. Explicit ties and conflicting decisions count as ties. A JoLT win, tie, or loss scores $1$, $0.5$, or $0$ for the preference score; decisive votes use two-sided exact sign tests with Holm correction.

\subsection{Extended Complexity Metrics}
\label{sec:appendix:extended-complexity}

\Cref{tab:extended-complexity} complements the three primary complexity metrics of \cref{tab:main} with five additional texture and information measures. DCT and JPEG ratio are measured after matched $1024^2$ px rescaling so these resolution-sensitive measures compare content at a common pixel budget rather than rewarding additional pixels. JPEG ratio is a compression-based complexity proxy rather than a direct image-quality measure. The main-text pattern persists: \METHODNAME{} leads GLCM contrast by $58\%$, edge density by $29\%$, and gradient energy by $47\%$ over the strongest baseline, and produces $2.4\times$ the Laplacian variance, all significant under the paired test. The sole exception is luminance entropy: all four methods land within half a bit of the 8-bit ceiling, so it cannot distinguish between them ($p_{\mathrm{adj}}=.95$). These results corroborate that the complexity gains in \cref{tab:main} are not an artifact of the three displayed metrics.

\begin{table}[t]
\centering
\caption{\textbf{Extended complexity metrics.} Five additional texture and information measures over the 200 evaluation prompts complement the three primary metrics in \cref{tab:main}. Entries are mean $\pm$ sample standard deviation; higher values indicate greater complexity rather than universally better quality. Boldface and underlining mark the best and second-best means only when their paired difference is significant after Holm correction within this table ($p_{\mathrm{adj}}<.05$, two-sided Wilcoxon). GLCM contrast is measured after the same matched $1024^2$ px rescaling as DCT and JPEG ratio in \cref{tab:main}. Luminance entropy is bounded above by 8 bits for 8-bit images and sits near that ceiling for every method, so its differences carry little information.}
\label{tab:extended-complexity}
\begingroup
\setlength{\tabcolsep}{3.0pt}
\renewcommand{\arraystretch}{1.08}
\resizebox{\linewidth}{!}{%
\begin{tabular}{lccccc}
\toprule
Method & GLCM ($\uparrow$) & Entropy ($\uparrow$) & Edge dens. ($\uparrow$) & Grad. energy ($\uparrow$) & Lap. var. ($\uparrow$) \\
\midrule
SEGA & $7.45\!\pm\!3.90$ & $7.568\!\pm\!0.248$ & $\underline{0.454\!\pm\!0.150}$ & $\underline{0.059\!\pm\!0.022}$ & $0.0012\!\pm\!0.0009$ \\
ResDiT & $6.84\!\pm\!3.29$ & $7.662\!\pm\!0.209$ & $0.422\!\pm\!0.131$ & $0.054\!\pm\!0.019$ & $0.0011\!\pm\!0.0008$ \\
DemoFusion & $\underline{9.02\!\pm\!3.77}$ & $7.744\!\pm\!0.263$ & $0.402\!\pm\!0.109$ & $0.055\!\pm\!0.018$ & $\underline{0.0013\!\pm\!0.0009}$ \\
\textbf{Ours} & $\mathbf{14.23\!\pm\!2.59}$ & $7.782\!\pm\!0.089$ & $\mathbf{0.584\!\pm\!0.078}$ & $\mathbf{0.087\!\pm\!0.013}$ & $\mathbf{0.0032\!\pm\!0.0008}$ \\
\bottomrule
\end{tabular}
}
\endgroup
\end{table}

\subsection{Extended Text-Alignment Metrics}
\label{sec:appendix:extended-alignment}

\Cref{tab:extended-alignment} reports the remaining text-alignment judges alongside the LMM4LMM correspondence score from \cref{tab:main}, and makes the complexity--alignment trade-off of \cref{sec:experiments:quantitative} explicit. The judges split cleanly by how they treat content beyond the prompt. Decompositional judges, which verify prompt elements individually, all rank \METHODNAME{} last, with gaps to the best baseline between $10\%$ (TIT-Score) and $43\%$ (TIT-Score-LLM). This group includes both DSG variants, VQAScore, both TIT scores, and QIB Alignment. The holistic LMM4LMM correspondence judgment instead places us within $0.004$ of the best method, a statistically indistinguishable gap. This split is the expected signature of our detail prompt: it deliberately adds locally plausible objects and textures beyond the master prompt, and the resulting density can crowd prompted elements and make individual element checks fail (see also the readability hypothesis in \cref{sec:limitations_and_conclusion}), whereas a holistic correspondence judgment remains at baseline level. The user study corroborates the holistic reading: in \cref{tab:user-study}, human raters prefer \METHODNAME{} for prompt match over SEGA and ResDiT and show no significant preference against DemoFusion, a pattern the decompositional rankings contradict. No baseline dominates the decompositional judges either: ResDiT leads VQAScore while DemoFusion leads the DSG variants and both TIT scores. Consistent with this distinction, the component ablation in \cref{fig:component-ablation} uses the holistic LMM4LMM correspondence score and finds a $2.9\%$ increase when joint-prior conditioning replaces the static prior at the same resolution.

\begin{table}[t]
\centering
\caption{\textbf{Extended text-alignment metrics.} Decompositional and holistic prompt--image alignment scores over the 200 evaluation prompts. LMM4LMM correspondence is repeated from \cref{tab:main} for reference. Entries are mean $\pm$ sample standard deviation. Boldface and underlining mark the best and second-best means only when their paired difference is significant after Holm correction within this table ($p_{\mathrm{adj}}<.05$, two-sided Wilcoxon). DSG, VQAScore, TIT-Score, and TIT-Score-LLM are evaluated with the Qwen3.6-27B backbone. Both CLIPScore variants are excluded here because some evaluation prompts exceed the CLIP text encoder's 77-token window; they appear with the other context-limited metrics in \cref{tab:legacy-metrics}.}
\label{tab:extended-alignment}
\begingroup
\setlength{\tabcolsep}{3.0pt}
\renewcommand{\arraystretch}{1.08}
\resizebox{\linewidth}{!}{%
\begin{tabular}{lccccccc}
\toprule
Method & DSG ($\uparrow$) & DSG no-dep. ($\uparrow$) & VQAScore ($\uparrow$) & TIT ($\uparrow$) & TIT-LLM ($\uparrow$) & QIB Align. ($\uparrow$) & LMM corr. ($\uparrow$) \\
\midrule
SEGA & $0.509\!\pm\!0.153$ & $0.580\!\pm\!0.142$ & $0.718\!\pm\!0.286$ & $0.563\!\pm\!0.068$ & $0.520\!\pm\!0.211$ & $49.3\!\pm\!15.3$ & $0.371\!\pm\!0.038$ \\
ResDiT & $0.566\!\pm\!0.174$ & $\underline{0.630\!\pm\!0.152}$ & $\mathbf{0.892\!\pm\!0.171}$ & $\underline{0.571\!\pm\!0.062}$ & $\underline{0.572\!\pm\!0.181}$ & $53.0\!\pm\!16.4$ & $0.360\!\pm\!0.043$ \\
DemoFusion & $0.588\!\pm\!0.156$ & $\mathbf{0.657\!\pm\!0.134}$ & $\underline{0.840\!\pm\!0.245}$ & $\mathbf{0.583\!\pm\!0.060}$ & $\mathbf{0.623\!\pm\!0.177}$ & $53.1\!\pm\!13.1$ & $0.356\!\pm\!0.044$ \\
\textbf{Ours} & $0.463\!\pm\!0.153$ & $0.517\!\pm\!0.140$ & $0.628\!\pm\!0.322$ & $0.522\!\pm\!0.067$ & $0.352\!\pm\!0.189$ & $46.0\!\pm\!16.8$ & $0.367\!\pm\!0.047$ \\
\bottomrule
\end{tabular}
}
\endgroup
\end{table}

\subsection{Agreement Among Text-Alignment Metrics}
\label{sec:appendix:metric-agreement}

To interpret the disagreements in \cref{tab:extended-alignment}, we measure how strongly the nine text-alignment metrics of our evaluation campaign (the seven in \cref{tab:extended-alignment} plus both CLIPScore variants) agree with one another.

\begin{figure}[H]
  \centering
  \resizebox{.82\linewidth}{!}{\input{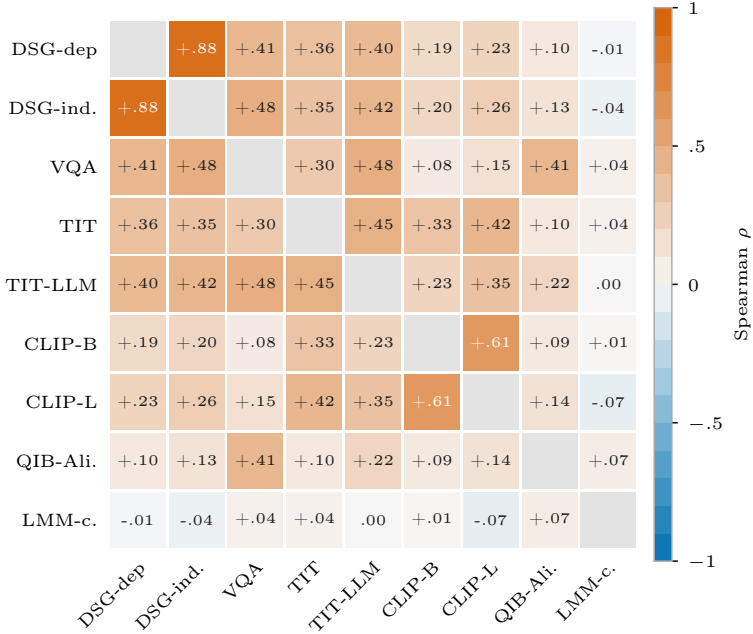}}
  \caption{\textbf{Pairwise agreement among the nine text-alignment metrics.} Spearman correlations over per-image scores pooled across the four compared methods and the four ablation configurations of \cref{fig:component-ablation} ($n=1{,}600$ images). The only strong correlations are within metric families: the two DSG variants and the two CLIPScore variants. Cross-family agreement is uniformly weak.}
  \label{fig:alignment-agreement}
\end{figure}

As \cref{fig:alignment-agreement} shows, the mean pairwise correlation is only $\rho=0.245$ (median $0.222$). The two strong pairs are within-family redundancies: the DSG variants at $\rho=0.88$ and the CLIPScore variants at $\rho=0.61$. No cross-family pair exceeds $\rho=0.48$, and the holistic LMM4LMM correspondence is essentially uncorrelated with every other judge ($|\rho|\leq0.08$). Centering scores within each prompt, which removes prompt difficulty as a shared driver, leaves this picture unchanged (mean $\rho=0.249$). These judges therefore measure substantially different constructs, and no single score is a definitive proxy for prompt fidelity in our regime; we read metric-specific alignment gaps as suggestive and rest our main alignment claims on the user study in \cref{sec:experiments:user-study}.

\subsection{Metric Profiles at a Glance}
\label{sec:appendix:radar}

\begin{figure}[H]
  \centering
  \resizebox{\linewidth}{!}{\input{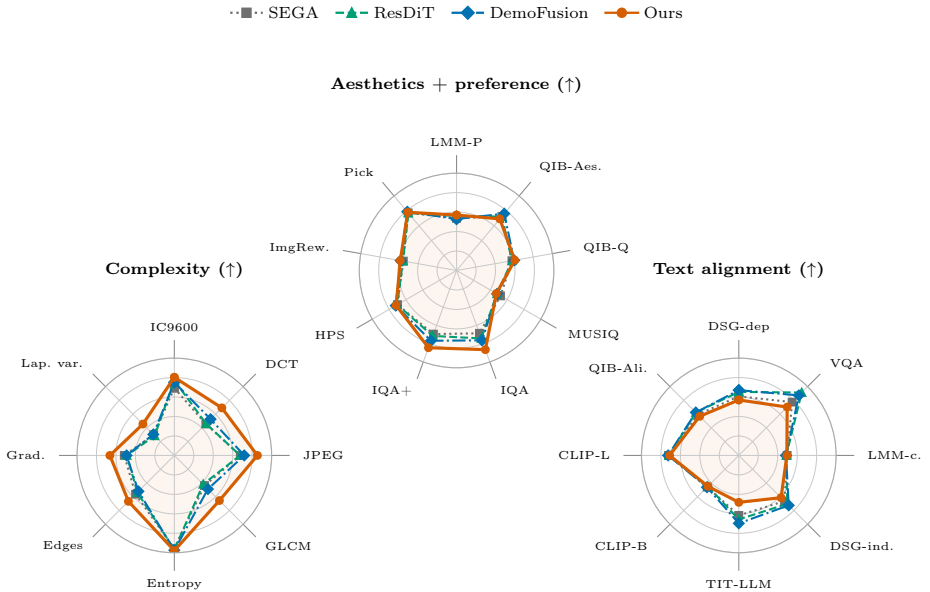}}
  \caption{\textbf{Metric profiles overview.} Each spoke is normalized to a domain fixed before plotting and independent of the four methods shown: calibrated or implementation-bounded metrics use their native reporting ranges, unbounded ones use endpoints rounded outward from the 99th percentile over all evaluated images, and learned rewards use fixed reporting windows. Outward means better (or more complex); radial magnitudes remain incomparable across spokes.}
  \label{fig:radar-appendix}
\end{figure}

\Cref{fig:radar-appendix} condenses \cref{tab:main,tab:preference} and the tables of this section into one view, without averaging axes into a single score. The complexity panel shows the uniform lead of \METHODNAME{} across all eight metrics; the text-alignment panel shows the decompositional--holistic split analyzed in \cref{sec:appendix:extended-alignment}; and the aesthetics-and-preference panel shows mixed rankings concentrated between \METHODNAME{} and DemoFusion. For instance, QIB Aesthetics favors DemoFusion ($70.3$ vs.\ our $61.5$), while LMM4LMM perception and QIB Quality favor our method. This mirrors the parity between the two methods in the user study (\cref{sec:experiments:user-study}).

\subsection{Legacy and Context-Limited Metrics}
\label{sec:appendix:legacy}

\begin{table}[H]
\centering
\captionsetup{font=footnotesize,skip=2pt}
\caption{\textbf{Legacy and sensitivity-only metrics.} These values are reported for comparison with prior work, not as primary evidence. For prompt-level metrics, boldface and underlining mark the best and second-best means only when their paired difference is significant after metric-family correction ($p_{\mathrm{adj}}<.05$, two-sided Wilcoxon). Inception Score (IS) is descriptive only and is left unstyled: it is computed at $299^2$ px, is not prompt-conditioned, and its reported deviation is across dataset splits rather than images.}
\label{tab:legacy-metrics}
\begingroup
\setlength{\tabcolsep}{3.0pt}
\renewcommand{\arraystretch}{1.0}
\resizebox{\linewidth}{!}{%
\begin{tabular}{lcccccc}
\toprule
Method & CLIP B/32 ($\uparrow$) & CLIP L/14 ($\uparrow$) & MUSIQ ($\uparrow$) & CLIP-IQA ($\uparrow$) & CLIP-IQA+ ($\uparrow$) & IS ($\uparrow$) \\
\midrule
SEGA & $\underline{0.810\!\pm\!0.068}$ & $\underline{30.51\!\pm\!3.61}$ & $\mathbf{40.19\!\pm\!14.25}$ & $0.608\!\pm\!0.089$ & $0.619\!\pm\!0.043$ & $5.48\!\pm\!0.83$ \\
ResDiT & $0.781\!\pm\!0.071$ & $29.62\!\pm\!3.69$ & $34.77\!\pm\!13.13$ & $0.679\!\pm\!0.077$ & $0.644\!\pm\!0.045$ & $4.80\!\pm\!0.59$ \\
DemoFusion & $\mathbf{0.820\!\pm\!0.077}$ & $\mathbf{31.64\!\pm\!4.00}$ & $\underline{35.15\!\pm\!12.62}$ & $\underline{0.704\!\pm\!0.067}$ & $\underline{0.711\!\pm\!0.043}$ & $5.42\!\pm\!0.83$ \\
\textbf{Ours} & $0.779\!\pm\!0.068$ & $27.92\!\pm\!3.39$ & $34.10\!\pm\!13.77$ & $\mathbf{0.832\!\pm\!0.027}$ & $\mathbf{0.806\!\pm\!0.021}$ & $4.36\!\pm\!0.55$ \\
\bottomrule
\end{tabular}
}
\endgroup
\end{table}

Some of the 200 semantically dense prompts exceed the 77-token context window of CLIP text encoders~\cite{radford2021learning}. CLIPScore~\cite{hessel2021clipscore} is therefore poorly matched to our evaluation setting: both B/32 and L/14 reduce the image to $224^2$ px and discard prompt content beyond that window.

MUSIQ~\cite{ke2021musiq} and CLIP-IQA(+)~\cite{wang2023exploring} were run on native $2048^2$ px and $4096^2$ px tensors, far beyond their calibrated resolutions, so their ordering is entangled with resolution and preprocessing. Inception Score~\cite{salimans2016improved} is dataset-level, not conditioned on the prompt, and unstable for small splits. \Cref{tab:legacy-metrics} therefore records these values for reproducibility only, outside the paper's main claims.

\end{document}